\documentclass[11pt]{article}

\usepackage[preprint]{acl}

\usepackage{times}
\usepackage{latexsym}

\usepackage[T1]{fontenc}

\usepackage[utf8]{inputenc}
\usepackage{amsmath}
\usepackage{amssymb}
\usepackage{multirow}
\usepackage{subcaption}
\usepackage{microtype}
\usepackage{makecell}

\usepackage{inconsolata}

\usepackage{graphicx}
\usepackage[table]{xcolor}
\usepackage{booktabs}

\title{From Candidate Tokens to Final Outcomes:  Decoding Relative Long-Horizon Candidate Value from Reasoning Model Representations}

\author{\textbf{{Yixin Yang}\textsuperscript{\rm 1}, {Yue Wang}\textsuperscript{\rm 1}, {Fandong Meng}\textsuperscript{\rm 2}, {Zhifang Sui}\textsuperscript{\rm 1}} \\
\textsuperscript{\rm 1} State Key Laboratory of Multimedia Information Processing, \\ School of Computer Science, Peking University \\
  \textsuperscript{\rm 2} WeChat AI, Tencent Inc., China \\ 
  \tt yangyx@stu.pku.edu.cn, wyy209@pku.edu.cn\\
  \tt fandongmeng@tencent.com, szf@pku.edu.cn}

\begin{document}
\maketitle
\begin{abstract}
Large reasoning models generate reasoning trajectories token by token. At a given prefix, multiple next-token candidates may be plausible, yet shape subsequent reasoning and differ in their chances of leading to the correct answer. Whether these differences are reflected in internal representations remains unclear. To study this question, we define relative long-horizon candidate value to quantify within-prefix differences and construct the Reasoning Candidate-Token Value (RCTV) dataset, pairing mathematical reasoning prefixes with plausible next-token candidates and their empirical long-horizon candidate values. Using RCTV, we examine hidden states, multilayer perceptron outputs, and attention outputs at candidate positions with sparse linear probes. Experimental results provide evidence that relative long-horizon candidate value is linearly decodable from these representations at multiple layers. 
Candidate-value probes can also outperform probability- and embedding-based baselines.
Furthermore, a controlled online intervention experiment shows that using the decoded signals to guide token selection during generation can improve the chances of eventual success.
Our findings provide evidence that internal representations encode information about relative long-horizon candidate value, offering insights into how local candidate-level representations relate to long-horizon reasoning outcomes.
\end{abstract}

\section{Introduction}
Large reasoning models tackle complex tasks, such as mathematical problem solving, by autoregressively generating extended chains of reasoning~\citep{guo2025deepseek,qwen3technicalreport,jiang2026think}. In this process, individual tokens can shape subsequent reasoning trajectories and final outcomes, with some exerting particularly consequential effects~\citep{bigelow2025forking,zur2025language,ko2026cliff}. Thus, although token choices are made locally, their consequences can extend over the remainder of the reasoning process. Yet autoregressive decoding is often regarded as inherently myopic~\citep{ma2025non, li2026martingale}. This tension raises the question of whether these long-horizon signals may nevertheless be present within the model.

\begin{figure*}[t]     
\centering     
\includegraphics[
  width=\linewidth,
  trim=0 16pt 0 10pt,
  clip
]{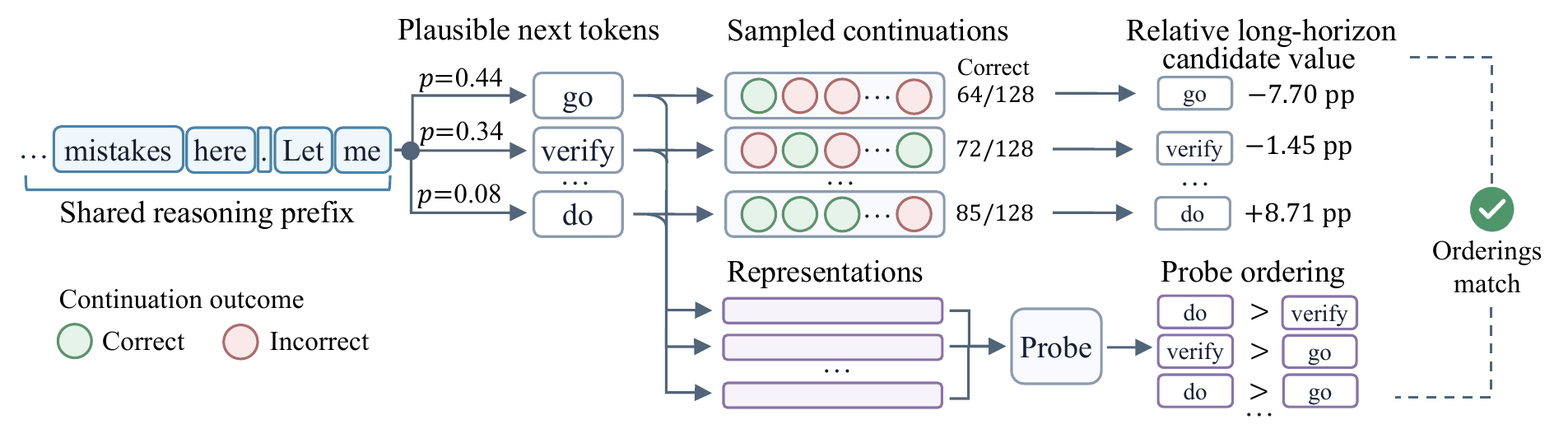}
\caption{
Illustration of our candidate-level analysis of long-horizon reasoning outcomes.
Three plausible candidates at a shared reasoning prefix are shown. \textit{pp} denotes percentage points.
In this example, the pairwise ordering produced by the probe matches the empirical ordering of candidates by relative long-horizon value.
}
\label{fig:rtcv-overview}
\end{figure*}

Internal representations provide a natural lens for examining outcome-relevant information in reasoning models. Prior work has found information about eventual success in representations extracted before generation~\citep{lugoloobi2026llms,choi2026your}, along partial reasoning trajectories~\citep{sun2026llm,yuan2026hidden}, and after generation ~\citep{liang2026your}, as well as information about outcome uncertainty \citep{zur2025language}.  These studies characterize outcome-related signals at the level of a prompt, an observed trajectory, or a position along it. At a given prefix, however, 
there are alternative next-token candidates that may differ substantially in their chances of leading to eventual success (Figure~\ref{fig:rtcv-overview}). Whether these differences are reflected in the model's candidate-related representations remains unclear.
If so, this would reveal an internal distinction among alternative token choices associated with different chances of eventual success, before their continuations are generated.
Such a finding would connect candidate-level representations to long-horizon reasoning outcomes and offer an internal view of alternatives available at the same prefix, including choices not taken in the observed trajectory. We therefore ask: \emph{Do reasoning models' internal representations distinguish plausible next-token candidates by their relative chances of eventual success?}

To study this question, we introduce relative long-horizon candidate value, a within-prefix measure of how plausible candidates differ in their chances of eventual success.
We then construct the Reasoning Candidate-Token Value (RCTV) dataset from mathematical reasoning problems with verifiable answers. For each prefix, we pair plausible next-token candidates with their empirical long-horizon values.
Using RCTV, we apply sparse linear probes to test whether relative long-horizon candidate value is decodable from candidate-position representations.
We probe hidden states, self-attention outputs, and multilayer perceptron (MLP) outputs at every model layer.
Finally, we use the trained probes to guide online token interventions as a functional test.

Experiments show that relative long-horizon candidate value is linearly decodable from internal representations. 
Probes on hidden states achieve higher pairwise AUC mainly in middle-to-late layers, and probes on MLP outputs achieve high pairwise AUC with only a few selected dimensions.
We also find that sparse linear probes can outperform probability-, rank- and embedding-based baselines.
Further analysis of MLP outputs shows that certain dimensions are repeatedly selected and exhibit consistent correlation signs across training folds.
We also observe above-chance probing performance in two additional reasoning models and in a different range of next-token entropy for the primary model.
Finally, controlled online interventions show gains in both continuation accuracy and prefix-conditioned Pass@\(k\) under probe guidance.

In summary, our contributions are as follows:
\begin{itemize}
    \item We define relative long-horizon candidate value and construct RCTV, a dataset pairing reasoning prefixes with plausible token candidates and their empirical long-horizon values.
    \item We show that relative long-horizon candidate value is linearly decodable from candidate-position representations across representation types.  We further extend the probing analysis across reasoning models and entropy regimes.
    \item We conduct a controlled functional test of probe-guided online token interventions, showing that such guidance can improve the chances of eventual success.
\end{itemize}

\section{Related Work}

\paragraph{Outcome-Relevant Information in Reasoning Model Representations}
Understanding whether internal representations of reasoning models contain outcome-relevant information has drawn increasing attention.
Prior work on prompt-level signals shows that pre-generation activations or initial hidden states are informative about eventual task success~\citep{lugoloobi2026llms}, model-perceived difficulty~\citep{zhu2025llm}, or expected task performance~\citep{choi2026your}.
Other work turns to partial reasoning trajectories, showing that representations from early or incomplete CoT prefixes already reflect outcome-relevant signals, including eventual reasoning success~\citep{afzal2025knowing,yuan2026hidden,sun2026llm,xu2026sparse}, upcoming intermediate-answer correctness~\citep{zhang2504reasoning}, and emerging final-answer beliefs~\citep{boppana2026reasoning}.
Further work focuses on post-generation verification, using activation deltas~\citep{liang2026your} and trajectory-level geometry~\citep{damirchi2026truth} from completed reasoning traces to distinguish correct from incorrect solutions.
Beyond these lines of work, recent work~\citep{zur2025language} has examined outcome uncertainty across token positions in a reasoning trajectory.
These studies characterize outcome-relevant information at the prompt, partial-trajectory, full-trajectory, or position level, rather than at the level of alternative next-token candidates available at the same prefix.

\paragraph{Token-Level Effects and Value in Reasoning}
Reasoning models generate solutions autoregressively, with each token becoming context for all subsequent generation.
Prior work~\citep{bigelow2025forking,zur2025language,wang1939beyond} shows that token choices at critical or high-entropy positions can materially alter subsequent reasoning, final outcomes, and outcome uncertainty.
Other studies~\citep{lin2024critical,ko2026cliff} find that certain individual tokens can predispose the reasoning trajectory to failure.
Recent work~\citep{zhu2025chain} provides causal evidence that selected CoT tokens store and propagate intermediate results.
FlashTrace traces attribution from generated spans through intermediate reasoning tokens back to source inputs~\citep{pan2026towards}.
Further work uses outcome feedback or reference continuations for value-guided tree search~\citep{lee2025token}, decoding and routing policies~\citep{li2026beyond,yin2026efficient,li2026futurebridge}, and token-level RL updates~\citep{deng2026token,gao2026unleashing,wang2026contrastive}.
From a training-dynamics perspective, \citet{cheng2026cancellation} use counterfactual token values to analyze which tokens are reinforced by critic-free RL.
Together, these studies highlight the importance of token-level analysis for understanding and improving reasoning.
Our study complements these efforts by directly measuring within-prefix differences in candidate continuation success rates and examining their linear decodability from frozen model representations.

\begin{table*}[t]
\footnotesize
\centering

\vspace{-8pt}
\setlength{\tabcolsep}{5pt}
\renewcommand{\arraystretch}{1.05}
\newcommand{\numshift}[1]{#1\hspace{10pt}}
\begin{tabular}{lllrrr}
\toprule
Subset
& Model
& Entropy
& \# Problems
& \# Candidates
& \# Rollouts \\
\midrule
RCTV-Core & Qwen3-1.7B & $[q_{.95}, 1]$ & \numshift{500} & \numshift{2,715} &  347,520 \\
RCTV-Qwen8B & Qwen3-8B & $[q_{.95}, 1]$ & \numshift{500} & \numshift{3,152}  & 403,456 \\
RCTV-Llama8B & DS-Llama-3.1-8B &  $[q_{.95}, 1]$ & \numshift{200} &  \numshift{3,051} & 390,528 \\
RCTV-InterEnt & Qwen3-1.7B &  $[q_{.85}, q_{.95})$ & \numshift{500} & \numshift{1,678} &  214,784 \\
\bottomrule
\end{tabular}
\caption{Dataset statistics for RCTV. The four subsets cover three reasoning models and two entropy regimes.}
\label{tab:rtcv-stats}
\end{table*}

\section{Measuring the Long-Horizon Value of Candidate Tokens}

Reasoning trajectories are generated token by token. 
The selected token shapes the subsequent trajectory and may affect the final answer.
In this section, we formalize the long-horizon value of candidate tokens and introduce Reasoning Candidate-Token Value (RCTV), a dataset for probing candidate-level outcome signals in model representations. The overall setup is illustrated in Figure~\ref{fig:rtcv-overview}.

\subsection{Relative Long-Horizon Candidate Value}
\label{sec:relative-value}
Given a reasoning prefix \(x\), let \(\mathcal{C}(x)\) denote the set of plausible next-token candidates under the policy \(\pi_\theta(\cdot \mid x)\). 
Each candidate \(t\in\mathcal{C}(x)\) is assigned a long-horizon candidate value \(v(x,t)\), defined as the expected final-answer correctness when generation continues from the prefix \(x\oplus t\), where \(\oplus\) denotes token concatenation. 
Because absolute values are affected by both problem difficulty and properties of the prefix, our analysis uses relative long-horizon candidate values within each plausible candidate set. 
This within-prefix comparison holds the reasoning context fixed and focuses the analysis on differences among candidate tokens.
Specifically, the mean of all plausible candidates serves as a prefix-specific reference.
The relative long-horizon candidate value is defined as
\begin{equation}
\label{eq:relative-value}
\Delta v(x,t)
=
v(x,t)
-
\frac{1}{|\mathcal{C}(x)|}
\sum_{t'\in\mathcal{C}(x)} v(x,t').
\end{equation}
Direct observation of the long-horizon value \(v(x,t)\) is infeasible. In practice, we estimate it with Monte Carlo rollouts by appending candidate $t$ to the prefix and sampling $K$ independent continuations from the same policy.
Given \(K\) sampled continuations \(\{y^{(j)}\}_{j=1}^{K}\sim \pi_\theta(\cdot \mid x\oplus t)\), a reference answer \(a^\star\), and an answer-extraction function \(\mathrm{Ans}(\cdot)\), the empirical long-horizon value is defined as
\begin{equation}
\label{eq:empirical-value}
\hat v(x,t)
=
\frac{1}{K}
\sum_{j=1}^{K}
\mathbf{1}\!\left[
\mathrm{Ans}(y^{(j)}) = a^\star
\right].
\end{equation}
The empirical target used in our analyses is the plug-in estimate \(\Delta \hat v(x,t)\), obtained by substituting \(\hat v\) for \(v\) in Eq.~\ref{eq:relative-value}. 
Hereafter, we use \emph{candidate value} and \emph{relative candidate value} as shorthand for \emph{long-horizon candidate value} and \emph{relative long-horizon candidate value}, respectively.

\subsection{Dataset Construction: Reasoning Candidate-Token Value}
\label{sec:data_construction}
We construct the Reasoning Candidate-Token Value (RCTV) dataset to study differences in long-horizon candidate value among plausible next-token candidates during reasoning. 
Each sample contains a reasoning problem, a prefix, a set of plausible candidate tokens, and their long-horizon candidate values.  
The prefix, candidates, and candidate values are all obtained from the target reasoning model whose representations we later analyze.
The relative candidate values used in our analyses are derived from these candidate values as defined in Section~\ref{sec:relative-value}.
We use the Big-Math Dataset~\citep{albalak2025big} as our source problem pool. 
This dataset provides a large corpus of open-ended mathematical problems from diverse sources, each with a closed-form, automatically verifiable answer.

\textbf{(1) Problem and trajectory selection.}
Problems that are too easy or too difficult tend to yield saturated candidate values, where most candidate tokens lead to nearly always correct or incorrect continuations. This leaves little room for meaningful within-prefix comparison. 
We therefore focus on problems of intermediate difficulty. 
For each problem, we generate eight solutions with the target model and retain problems solved correctly in two to six attempts.
We collect an equal number of problems from each selected success-count stratum and sample one source trajectory per problem.

\textbf{(2) Prefix construction.}
For each problem, we sample a position within its source reasoning trajectory according to its next-token entropy and take all preceding tokens as the corresponding prefix \(x_i\). Our primary subsets use positions satisfying \(H_{i,\mathrm{pos}}\geq q_{.95}\), where the entropy quantiles are computed from the empirical distribution across recorded positions. 
We additionally construct RCTV-InterEnt using positions in the intermediate entropy range \(q_{.85}\leq H_{i,\mathrm{pos}}<q_{.95}\). 
Full details of position sampling are in
Appendix~\ref{app:rtcv-position-selection}. 

\textbf{(3) Candidate selection and rollout collection.}
At each problem and its selected prefix, we first identify plausible candidates from the model's next-token distribution. 
Specifically, we define \(\mathcal{C}(x_i)=\{t\in\mathcal{V}:
p_\theta(t\mid x_i)\geq\tau_{\mathrm{cand}}\}\), where \(\mathcal{V}\) denotes the model vocabulary, and set \(\tau_{\mathrm{cand}}=0.01\) for all RCTV subsets. 
For each retained candidate \(t\), we sample 128 independent
continuations conditioned on \(x\oplus t\). 
As defined in Section~\ref{sec:relative-value}, the fraction of verified correct continuations yields \(\hat v(x_i,t)\), from which \(\Delta\hat v(x_i,t)\) is derived.
Detailed generation and verification settings are in Appendix~\ref{app:generation-verification}.

\textbf{Summary statistics.} We construct four variants of RCTV, covering two entropy thresholds and three models across two model families. RCTV-Core serves as our primary dataset for our main experiments. RCTV-InterEnt varies the entropy range, while RCTV-Qwen8B and RCTV-Llama8B vary the generation model. In total, the dataset comprises 1,400 samples across all variants. Full statistics for each variant are reported in Table~\ref{tab:rtcv-stats}. We further quantify the within-prefix value gaps in RCTV-Core. For each problem, the value gap is defined as the difference between the maximum and minimum long-horizon value among its candidate tokens. 
Appendix~Figure~\ref{fig:value-gap-distribution} shows the distribution of value gaps, with a mean of 7.3 percentage points and a 90th percentile of 14.1 percentage points.
See Appendix~\ref{app:RCTV_Statistics} for additional statistics and checks on rollout-based value estimation.

\section{Decoding Candidate Value from Internal Representations}
The consequences of a candidate next token may not become apparent until the remainder of the reasoning trajectory has unfolded. As shown in the previous section, candidate tokens from the same prefix can differ in their long-horizon value. In this section, we ask whether these relative value differences are reflected in the model's internal representations extracted before the subsequent reasoning trajectory is generated.

\subsection{Probing Setup}
\textbf{Model.} We keep the parameters of all reasoning models frozen throughout probing and update only the probe parameters. For each model, we train and evaluate the probes on its corresponding RCTV subset. We conduct the primary analysis with Qwen3-1.7B~\citep{qwen3technicalreport} and repeat the probing analysis with Qwen3-8B and DeepSeek-R1-Distill-Llama-8B (DS-Llama-3.1-8B; \citealp{deepseekai2025deepseekr1incentivizingreasoningcapability}). Detailed model specifications are provided in Appendix~\ref{app:model-specifications}.

\textbf{Representations.}
We consider three types of internal representations at each decoder layer:
(1) the hidden state, given by the complete output of the decoder layer;
(2) the self-attention output, given by the residual update produced by the self-attention sublayer after its output projection; and
(3) the multilayer perceptron (MLP) output, given by the residual update produced by the MLP sublayer after its final projection. The latter two are extracted before they are added to the residual stream.
For each prefix \(x\) and candidate token \(t\), our representation is extracted at the candidate-token position after appending \(t\) to \(x\).
Let \(\mathbf{R}_{\ell,r}(s)_j\) denote the representation of type \(r\) at layer \(\ell\) and token position \(j\), obtained when the model processes sequence \(s\).
We denote this candidate-position representation conditioned on the shared prefix by
\begin{equation}
\mathbf{z}_{\ell,r}(x,t)
=
\mathbf{R}_{\ell,r}\!\left(x \oplus t\right)_{|x|+1}.
\label{eq:candidate-position-representation}
\end{equation}
Additionally, we examine prefix--token interaction representations constructed before \(t\) is processed by the
model in Appendix~\ref{app:prefix-token-interactions}.

\textbf{Candidate-value readout.}
Following prior work on sparse probing
~\citep{gurnee2023finding,alabi2024hidden}, we use correlation-ranked
top-\(k\) linear probes as candidate-value
readouts to predict the relative long-horizon candidate value
\(\Delta \hat v(x,t)\) from each candidate representation \(\mathbf{z}_{\ell,r}(x,t)\). 
Within each fold, we rank representation dimensions by their absolute Pearson correlation with \(\Delta \hat v(x,t)\), computed exclusively on the training portion, and retain the top \(k\) dimensions.
We then fit a ridge-regularized linear probe on the selected dimensions to predict \(\Delta \hat v(x,t)\).
Each candidate token is scored only when its associated sample belongs to the held-out fold. We aggregate the held-out candidate scores across the five folds to obtain out-of-fold predictions for all candidates, on which all evaluation metrics are computed.
The complete feature-selection and optimization procedure is provided in
Appendix~\ref{app:sparse-readout}.
For single-site representations, we evaluate \(k\in\{2,4,8,16,32,64,128,256\}\). 

\begin{figure*}[!tp]
    \centering
    \includegraphics[width=\textwidth]{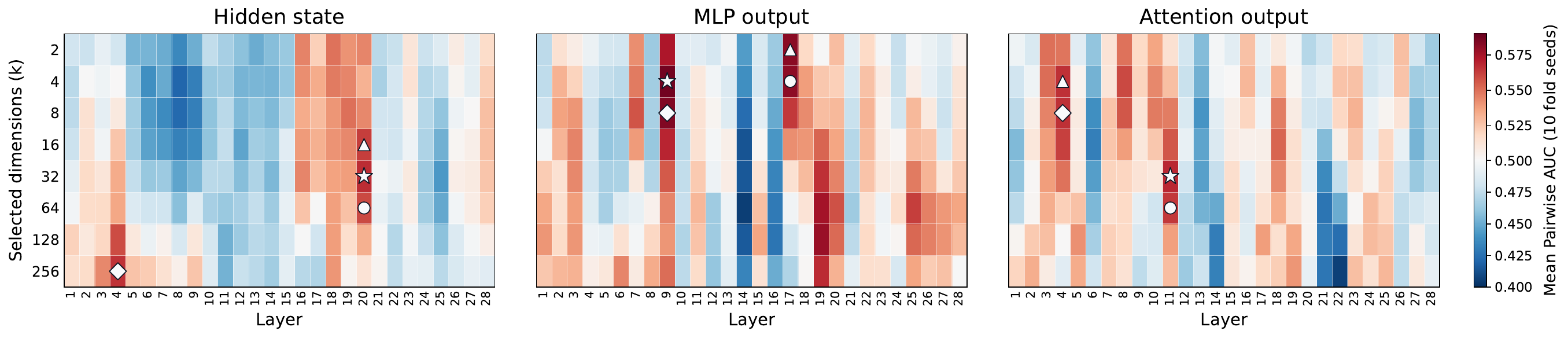}
    \caption{
    Candidate-value readout across layers and selected dimensionalities.
    Each cell reports the mean pairwise AUC over ten problem-level fold seeds
    for hidden states, MLP outputs, and attention outputs.
    Within each representation family, $\star$, $\diamond$, $\triangle$, and
    $\circ$ denote the first- through fourth-highest observed mean AUC,
    respectively. 
    }
    \label{fig:candidate_position_readout_heatmap}
\end{figure*}

\subsection{Evaluation Metrics}
For each prefix, the probe assigns a score to each candidate token.
The evaluation therefore focuses on two aspects of candidate comparison.
Pairwise AUC and tier AUC measure agreement between probe scores and relative candidate values, whereas Regret@1 measures the quality of the top-scoring candidate.

\textbf{Pairwise AUC.}
Pairwise AUC evaluates the ordering of candidate tokens at the same prefix. To exclude near-ties, we retain only pairs satisfying \(\lvert \Delta\hat v(x,t_i)-\Delta\hat v(x,t_j)\rvert\ge m\), where \(m\) is the value-gap margin. 
For each retained pair, we define the label as \(y_{ij}=\mathbb{I}[\Delta\hat v(x,t_i)>\Delta\hat v(x,t_j)]\) and the pairwise probe score as \(s_{ij}=g(x,t_i)-g(x,t_j)\), where \(g(x,t)\) is the out-of-fold probe prediction. 
Pairwise AUC is the ROC AUC computed from \(y_{ij}\) and \(s_{ij}\) over the within-prefix pairs.

\textbf{Tier AUC.} Tier AUC measures discrimination between candidates in the positive and negative relative-value tiers.
We assign candidate tokens with \(\Delta \hat v(x,t)>\epsilon\) to the positive tier
and candidates with \(\Delta \hat v(x,t)<-\epsilon\) to the negative tier.
Candidates with \(-\epsilon\le\Delta \hat v(x,t)\le\epsilon\) are excluded to
avoid treating small deviations around the prefix mean as reliable tier
assignments.
Before evaluation, we center the probe scores within each prefix by subtracting their mean over its candidate tokens.
Tier AUC is the ROC AUC computed from these centered scores and the positive-versus-negative tier labels over all retained candidates. 

\textbf{Regret@1.}
For each prefix \(x\), let \(t_{\mathrm{sel}}(x)\) denote the candidate assigned the highest probe score. We define
\begin{equation}
\mathrm{Regret@1}(x)
=
\max_{t\in\mathcal{C}(x)}\hat v(x,t)
-
\hat v(x,t_{\mathrm{sel}}(x)).
\label{eq:regret-at-one}
\end{equation}
We report mean Regret@1 across prefixes. 
Regret@1 measures the loss in candidate value incurred by the probe's top selection.
Lower is better, with zero indicating that the selected candidate attains the maximum candidate value among \(\mathcal{C}(x)\). 

We use pairwise AUC as the primary metric. Higher pairwise AUC and tier AUC values are better, whereas lower Regret@1 values are better.
We use \(m=0.08\) and \(\epsilon=0.04\) as fixed thresholds in the main analysis and report additional statistics and results across other threshold values in Appendix~\ref{app:threshold-sensitivity}. 

\subsection{Baselines}
We compare candidate-value readouts from internal representations with five baselines. All baselines are evaluated using the same folds, candidate sets, and metrics as the candidate-value readouts. 
The five baselines are:
\emph{(1) Uniform random.}
This baseline selects a candidate uniformly at random from each candidate set. We report its selection metrics exactly in expectation, with pairwise and tier AUC equal to \(0.5\). 
\emph{(2) Next-token log-probability.}
We directly use the model's next-token log-probability, \(\log p_\theta(t\mid x)\), as the candidate score. This baseline assesses the extent to which candidate value is captured by the model's local next-token preference. 
\emph{(3) Probability rank.}
We use the negative within-prefix probability rank,
\(-\operatorname{rank}_{p_\theta}(t\mid x)\), as the candidate score,
so that larger scores correspond to more probable candidates.
This baseline evaluates whether probability ordering alone captures candidate value.
\emph{(4) Token embedding.}
We represent each candidate using only its input embedding
\(\mathbf{e}_t\) and apply the same correlation-ranked top-\(k\) ridge probe. This baseline tests whether candidate value can be predicted from the
candidate token identity alone.
\emph{(5) Activation norm.}
We replace the full activation vector with its Euclidean norm and train a one-dimensional ridge probe. This control tests whether activation magnitude alone predicts relative candidate value.

\subsection{Linear Decodability of Candidate Value}
We evaluate candidate-value readout from representations at the candidate-token position. We repeat five-fold cross-validation with ten problem-level fold seeds and examine hidden state, MLP output, and self-attention output representations across layers and numbers of selected dimensions. 

\textbf{Candidate-value signal is unevenly distributed across layers and
dimensions, with different patterns across representation families.}
Figure~\ref{fig:candidate_position_readout_heatmap} shows the mean pairwise AUC over ten
problem-level fold seeds for each layer and number of selected dimensions.
Hidden states show higher pairwise AUC mainly in middle-to-late layers, with a weaker signal in early layers at high dimensionality.
MLP outputs exhibit peaks with very few selected dimensions, particularly near layers 9 and 17.
Attention outputs show more scattered regions of higher pairwise AUC. Appendix~\ref{app:readout-heatmaps} further shows heatmaps of standard deviation
and above-chance frequency across fold seeds.

\begin{table}[!htbp]
\small
\setlength{\tabcolsep}{2.8pt}
\begin{tabular}{lrccc}
\toprule
Repr
& $k$
& Pairwise $\uparrow$
& Tier $\uparrow$
& Regret@1 $\downarrow$\\
\midrule

\rowcolor{black!7}
\multicolumn{5}{l}{\textit{Baselines}} \\[2pt]
Uniform
& -
& 50.0
& 50.0
& 0.0355 \\

Log prob
& 1
& 50.9
& 51.6
& 0.0346 \\

Prob rank
& 1
& 52.4
& 51.8
& 0.0346 \\

Embedding
& 16
& 52.2 {\scriptsize$\pm$ 2.0}
& 52.2 {\scriptsize$\pm$ 1.9}
& 0.0340 {\scriptsize$\pm$ 0.0014} \\

Act norm ($\mathrm{M}_9$)
& 1
& 53.2 {\scriptsize$\pm$ 0.9}
& 54.5 {\scriptsize$\pm$ 0.8}
& 0.0338 {\scriptsize$\pm$ 0.0000}\\

\midrule
\rowcolor{black!7}
\multicolumn{5}{l}{\textit{Candidate-value readouts}} \\[2pt]

Hidden ($\mathrm{H}_{20}$)
& 32
& 56.6 {\scriptsize$\pm$ 2.3}
& \textbf{57.0} {\scriptsize$\pm$ 2.0}
& 0.0339 {\scriptsize$\pm$0.0010} \\

MLP ($\mathrm{M}_9$)
& 4
& \textbf{58.7} {\scriptsize$\pm$ 2.6}
& 56.3 {\scriptsize$\pm$ 2.5}
& \textbf{0.0323} {\scriptsize$\pm$ 0.0011}\\

Attn ($\mathrm{A}_{11}$)
& 32
& 56.8 {\scriptsize$\pm$ 2.3}
& 53.2 {\scriptsize$\pm$ 2.1}
& 0.0335 {\scriptsize$\pm$ 0.0013} \\

\bottomrule
\end{tabular}
\caption{
Candidate-value readout results and baselines.
We report the mean and standard deviation over ten problem-level fold seeds.
For each internal-representation family and the embedding baseline, we report the strongest observed configuration according to mean pairwise AUC.
}
\label{tab:candidate_value_readout}
\end{table}

\textbf{Candidate-value readout can outperform probability-, rank- and embedding-based baselines.} Table~\ref{tab:candidate_value_readout} reports the strongest observed
configuration in each representation family according to mean pairwise
AUC. All three families achieve higher pairwise and tier AUC and lower
Regret@1 than the probability, rank, and embedding baselines.
The MLP readout achieves the highest pairwise AUC and lowest Regret@1,
while the hidden-state readout achieves the highest tier AUC.
The MLP readout also outperforms its matched activation-norm baseline across all three metrics.
Taken together, the heatmaps and results across the three metrics provide evidence that \textbf{candidate value is linearly decodable from candidate-position representations}.

\begin{figure*}[!tp]
    \centering
    \includegraphics[width=\textwidth]
    {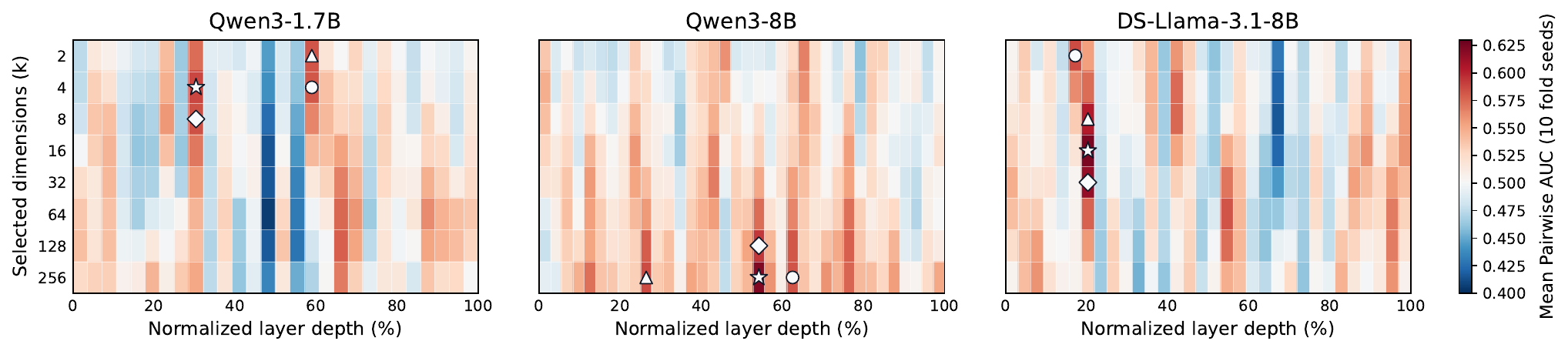}
    \caption{
    Candidate-value readout from candidate-position MLP outputs across
    Qwen3-1.7B, Qwen3-8B, and DS-Llama-3.1-8B.
    Each cell reports mean pairwise AUC over ten problem-level fold seeds.
    $\star$, $\diamond$, $\triangle$, and $\circ$ mark the first- through
    fourth-highest observed mean AUC within each model.
    Layer indices are normalized by model depth. 
    }
    \label{fig:cross_model_mlp_readout}
\end{figure*}

\begin{table*}[!htbp]
\centering
\small
\setlength{\tabcolsep}{6pt}
\begin{tabular}{cc c cc cc}
\toprule
\multicolumn{2}{c}{Configuration}
& \multicolumn{1}{c}{Main readout}
& \multicolumn{2}{c}{Controls}
& \multicolumn{2}{c}{Alternative readouts} \\
\cmidrule(lr){1-2}
\cmidrule(lr){3-3}
\cmidrule(lr){4-5}
\cmidrule(lr){6-7}
Layer & $k$
& Top-$k$ ridge
& Random dims
& Permuted labels
& Signed sum
& Corr-weighted \\
\midrule
$\mathrm{M}_{9}$ & 4
& 58.7 {\scriptsize$\pm$ 2.6}
& 50.1 {\scriptsize$\pm$ 0.7}
& 49.9 {\scriptsize$\pm$ 0.7}
& 58.7 {\scriptsize$\pm$ 2.6}
& 58.8 {\scriptsize$\pm$ 2.5} \\

$\mathrm{M}_{9}$ & 8
& 58.3 {\scriptsize$\pm$ 2.3}
& 50.4 {\scriptsize$\pm$ 1.0}
& 50.5 {\scriptsize$\pm$ 0.9}
& 58.3 {\scriptsize$\pm$ 2.6}
& 58.7 {\scriptsize$\pm$ 2.4} \\

$\mathrm{M}_{17}$ & 2
& 58.1 {\scriptsize$\pm$ 2.5}
& 50.1 {\scriptsize$\pm$ 0.9}
& 49.6 {\scriptsize$\pm$ 0.5}
& 58.1 {\scriptsize$\pm$ 2.7}
& 58.3 {\scriptsize$\pm$ 2.6} \\

$\mathrm{M}_{17}$ & 4
& 58.0 {\scriptsize$\pm$ 2.7}
& 49.7 {\scriptsize$\pm$ 0.7}
& 49.6 {\scriptsize$\pm$ 0.6}
& 57.8 {\scriptsize$\pm$ 3.0}
& 58.2 {\scriptsize$\pm$ 2.8} \\
\bottomrule
\end{tabular}
\caption{
Robustness of four high-performing candidate-value MLP
readouts. Values are mean pairwise AUC (\%) and standard deviation over
ten problem-level fold seeds. The table compares the primary Top-$k$ ridge
readout with null controls and alternative readout rules.
Control repetitions are first averaged within each fold seed.
}
\label{tab:mlp_readout_robustness}
\end{table*}

\begin{table}[!htbp]
\centering
\footnotesize
\setlength{\tabcolsep}{3.0pt}

\begin{tabular}[t]{rcc}
\toprule
\multicolumn{3}{c}{$\mathrm{M}_{9}$ ($k=4$)} \\
\cmidrule(lr){1-3}
Dim & Selected & Direction \\
\midrule
432  & 45/50 & $+$ 45/45 \\
950  & 38/50 & $-$ 38/38 \\
256  & 18/50 & $+$ 18/18 \\
1274 & 17/50 & $-$ 17/17 \\
\bottomrule
\end{tabular}
\hspace{10pt}
\begin{tabular}[t]{rcc}
\toprule
\multicolumn{3}{c}{$\mathrm{M}_{17}$ ($k=4$)} \\
\cmidrule(lr){1-3}
Dim & Selected & Direction \\
\midrule
1834 & 47/50 & $+$ 47/47 \\
312  & 31/50 & $+$ 31/31 \\
2021 & 29/50 & $-$ 29/29 \\
822  & 16/50 & $-$ 16/16 \\
\bottomrule
\end{tabular}

\caption{
Selection frequency and sign consistency of the most frequently selected MLP dimensions across 10 problem-level fold seeds (50 training folds). Direction denotes the sign of the training-fold correlation. 
}
\label{tab:mlp_dimension_stability}
\end{table}

\subsection{Further Analysis of Candidate Value in MLP Outputs}
For further analysis, we focus on MLP outputs, which achieve the highest pairwise AUC and lowest Regret@1 in the main comparison. We examine four high-performing configurations marked on the heatmap: \(\mathrm{M}_9\) with \(k=4,8\) and \(\mathrm{M}_{17}\) with \(k=2,4\).

\textbf{Control experiments yield near-chance performance.} 
We apply random-dimension and permuted-label controls to the four candidate-value readouts.
The former replaces correlation-based dimension selection with uniform random selection, while the latter permutes candidate values within each training prefix.
For each of the ten fold seeds, we repeat each control 20 times, running the full five-fold procedure in each repetition to compute pairwise AUC.
As shown in Table~\ref{tab:mlp_readout_robustness}, both controls yield pairwise AUC
close to 50\% across all four configurations, whereas the
candidate-value readouts achieve approximately 58\%.

\textbf{Performance is comparable across readout variants.}
We compare ridge with signed-sum and correlation-weighted readouts
for the four configurations, as shown in Table~\ref{tab:mlp_readout_robustness}.
Signed-sum sums the selected features using the signs of their training-fold correlations with candidate value, while correlation-weighted readout additionally weights them by the correlation magnitudes.
Both achieve pairwise AUC comparable to ridge, suggesting that the observed performance does not depend on ridge fitting.

\textbf{Frequently selected dimensions show consistent correlation signs.}
Table~\ref{tab:mlp_dimension_stability} summarizes dimension selection for \(k=4\) across ten fold seeds, comprising 50 training folds.
Several dimensions recur across fold assignments: dimension 432 at \(\mathrm{M}_9\) is selected in 45 of 50 folds, and dimension 1834 at \(\mathrm{M}_{17}\) in 47.
The frequently selected dimensions have consistent training-fold correlation signs whenever selected.


\subsection{Linear Decodability Across Models and Entropy Regimes}
We next examine candidate-value readout in two additional models and across entropy regimes. Following the preceding MLP analysis, we focus on MLP outputs and use ten problem-level fold seeds.

\textbf{Candidate value is linearly decodable across models.}
We apply the same probing procedure to Qwen3-8B and
DS-Llama-3.1-8B.
Figure~\ref{fig:cross_model_mlp_readout} shows the mean pairwise AUC across layers and numbers of selected dimensions.
Both Qwen models show regions of higher pairwise AUC around 25--30\% and 55--65\% of normalized layer depth, with the strongest readouts in Qwen3-8B occurring at larger \(k\).
DS-Llama-3.1-8B instead shows a peak in early layers.
These results show that candidate-value readout performance varies across models. Additional results are provided in Appendix~\ref{app:cross-model-representations}.

\textbf{Candidate value is also linearly decodable in the
intermediate-entropy regime.}
We evaluate fixed dimensions identified on RCTV-Core using five-fold cross-validation within RCTV-InterEnt over ten problem-level fold seeds.
At \(\mathrm{M}_{17}\), the two dimensions 1834 and 312 support above-chance discrimination across all three readouts.
Ridge, signed-sum, and correlation-weighted readouts achieve mean pairwise AUCs of 57.3\%, 60.5\%, and 58.7\%, respectively, with corresponding tier AUCs of 56.1\%, 58.4\%, and 57.2\%.
Additional readout results are provided in Appendix~\ref{app:entropy-analysis}.

\subsection{Probe-Guided Online Token Interventions}
We use probe-guided online token interventions as a functional test of candidate-value readouts. We examine whether selecting candidates with higher readout scores improves continuation outcomes during generation.

\textbf{Intervention setup.}
For a controlled comparison, we first follow standard sampling to produce the same initial prefix for all strategies. Intervention strategies then selects candidates at eligible high-entropy positions, using shared sampling randomness throughout generation. 
For probe-guided intervention, we use \(\mathrm{M}_{17}\) with \(k=4\), one of the high-performing sparse configurations identified in the offline analysis.
At each intervention, the readout scores the candidates, and we greedily select the top-scoring one.
We compare probe-guided intervention with three strategies: top-probability selection, uniform selection from the same plausible-candidate set, and standard sampling without intervention.
We use candidate samples from 400 RCTV-Core problems for dimension selection and ridge-probe fitting. The remaining 100 problems are excluded from both steps and used to generate new trajectories for online intervention evaluation.
We evaluate intervention budgets \(B\in\{8,16,32,64\}\), where \(B\) is the maximum number of interventions per trajectory.
For each trajectory, we generate a prefix under the corresponding strategy up to a fixed length, then sample 128 continuations from that prefix.
We report mean continuation accuracy (Avg@128), the fraction of continuations yielding a correct final answer, and prefix-conditioned Pass@\(k\), which estimates the probability that at least one of \(k\) sampled continuations is correct, following~\citet{chen2021evaluating}.
Both metrics are averaged across the 100 evaluation problems.
Further details are in Appendix~\ref{app:online-interventions_setup}.

\begin{figure}[t]
    \centering
    \includegraphics[
        width=\linewidth
    ]{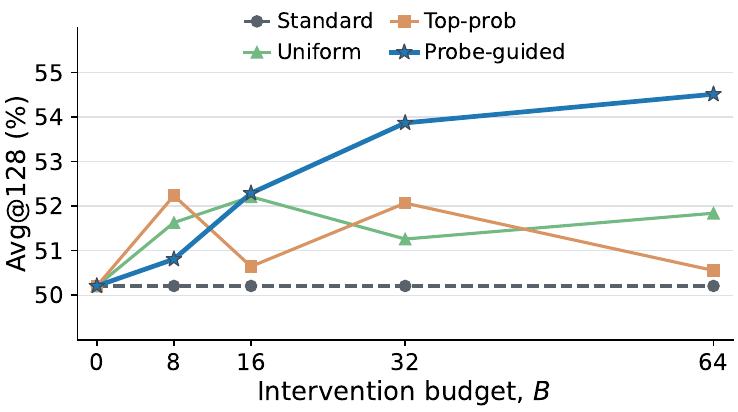}
    \caption{
    Mean continuation accuracy under different intervention budgets \(B\).
    The dashed line denotes generation without intervention.
    }
    \label{fig:online_accuracy_budget}
\end{figure}

\textbf{Intervention results.} 
Figure~\ref{fig:online_accuracy_budget} shows that probe-guided continuation accuracy increases with the intervention budget, reaching 54.51\% at \(B=64\).
As shown in Table~\ref{tab:online_b64}, this represents a gain of 4.30 percentage points over standard sampling, with a paired 95\% bootstrap confidence interval of [+0.60, +8.28].
Probe-guided interventions also achieve higher continuation accuracy than both top-probability and uniform selection.
The gains also extend to prefix-conditioned Pass@\(k\): probe-guided interventions achieve higher values than standard sampling across all evaluated budgets and
\(k\in\{4,8,16,32\}\), and exceed both other intervention strategies at \(B=32\) and \(B=64\).
Prefix-conditioned Pass@\(k\) results and additional intervention analyses are provided in Appendix~\ref{app:additional-intervention-results}.
These results suggest that offline probing can inform online token interventions that improve the chances of eventual success.

\begin{table}[t]
\centering
\small
\setlength{\tabcolsep}{3pt}
\begin{tabular}{lcc}
\toprule
\textbf{Policy} &
\textbf{Avg@128}  &
\textbf{\(\Delta\) vs. Standard (95\% CI)} \\
\midrule

Standard
& 50.21
& --- \\

Uniform
& 51.84
& +1.63 [-2.03, +5.30] \\

Top-prob
& 50.56
& +0.35 [-3.47, +4.18] \\

\textbf{Probe-guided}
& \textbf{54.51}
& \textbf{+4.30 [+0.60, +8.28]} \\

\bottomrule
\end{tabular}
\caption{
Online intervention results at \(B=64\).
Avg@128 is reported in percent; differences are in percentage points with 95\% paired problem-bootstrap CIs.
}
\label{tab:online_b64}
\end{table}

\section{Conclusion}
We investigate whether internal representations contain information about the relative long-horizon value of plausible next-token candidates at the same reasoning prefix.
We construct RCTV, which pairs mathematical reasoning prefixes with plausible candidates and  empirical relative candidate values.
Using RCTV, we find that relative candidate value is linearly decodable from candidate-position representations.
Experiments on additional models and an intermediate-entropy subset also show above-chance probing performance.
Beyond offline probing, controlled online interventions suggest that offline probing can inform token selection to improve the chances of eventual success.
Overall, our findings link candidate-level signals in internal representations to eventual reasoning outcomes.

\section*{Limitations}
Our experiments focus on mathematical reasoning problems with verifiable answers. Extending the analysis to other reasoning domains remains a direction for future work.
The online interventions serve as a controlled functional test of candidate-value readouts, rather than a general-purpose decoding method. We evaluate whether these readouts can inform token selection, while a systematic study of decoding efficiency and deployment is outside the scope of this work.
Our study examines candidate-value decodability and its use in online interventions; the underlying internal mechanisms could be explored in future work.


\bibliography{custom}

\appendix

\section{Additional RCTV Construction Details}
\label{app:rtcv-construction}

\subsection{Generation and Verification Settings}
\label{app:generation-verification}

All generations use a vanilla sampling configuration with temperature \(1.0\), top-\(p=1.0\), and a maximum generation length of 32K tokens. Generation terminates when the model produces an end-of-sequence token or reaches the maximum length.

To verify a continuation, we extract its final answer from the last valid \verb|\boxed{}| expression. We then use Math-Verify to determine whether the extracted answer is mathematically equivalent to the reference answer. A continuation is counted as correct only when an answer can be extracted and successfully verified as equivalent; otherwise, it is counted as incorrect.

\subsection{RCTV Position Selection}
\label{app:rtcv-position-selection}

We sample one source trajectory for each selected problem. For each trajectory, we select one token position as the candidate position, and the generated tokens preceding that position form the corresponding reasoning prefix. For computational efficiency, we record the next-token distribution only at the first 8,000 generated positions of each trajectory. 

We measure uncertainty at each recorded position using normalized top-10 entropy, computed by renormalizing the probabilities of the model's ten most probable next tokens and scaling the resulting entropy to \([0,1]\). The quantiles \(q_{.85}\), \(q_{.90}\), and \(q_{.95}\) are computed from entropy scores pooled across all recorded positions.
We exclude positions within the first 100 tokens of the reasoning segment to ensure that the prefix contains sufficient reasoning context, as well as positions within its final 100 tokens, where generation is typically dominated by answer finalization. 

For all RCTV subsets except RCTV-InterEnt, we uniformly sample from eligible positions with entropy at or above \(q_{.95}\). If a trajectory contains no such position, we instead sample from positions with entropy at or above \(q_{.90}\); if this set is also empty, we select the eligible position with the highest entropy in the trajectory. In RCTV-Core, only four of the 500 selected positions fall below \(q_{.95}\); the remaining 496 are drawn from the primary \([q_{.95},1]\) range. No fallback was required for RCTV-Qwen8B or RCTV-Llama8B.

RCTV-InterEnt instead targets the intermediate-entropy range \([q_{.85},q_{.95})\). For each of the same 500 source trajectories used in RCTV-Core, we uniformly sample one qualifying position. Candidate selection, continuation sampling, and value estimation otherwise follow the same procedure across all RCTV subsets.

\section{Additional RCTV Statistics and analyses}
\label{app:RCTV_Statistics}
\textbf{Additional dataset statistics.}
Table~\ref{tab:rtcv-additional-statistics} summarizes the sampled source trajectories, problem difficulties, prefix positions, and candidate-set sizes. Correct and incorrect source trajectories are approximately balanced in every variant. 
Each RCTV variant has an approximately uniform distribution across difficulty strata 2--6, where a problem's stratum is determined by the number of successful outcomes among eight sampled generations. We use eight generations to balance estimation reliability with the computational cost of dataset filtering. See Section~\ref{sec:data_construction} for details.
Across all variants, the selected prefixes span a wide range within the first 8,000 generated tokens rather than concentrating in a narrow region: the 10th percentiles fall between 420 and 851 tokens, while the 90th percentiles fall between 5,839 and 6,860 tokens.
Candidate-set sizes also vary across variants, reflecting differences in the models' next-token distributions and the sampled entropy ranges.

\begin{table*}[t]
    \caption{
        Additional statistics for the RCTV variants.
        C/W denotes prefixes drawn from source trajectories with correct and
        incorrect final answers.
        Difficulty entries report the numbers of problems in strata 2--6, where each stratum is determined by the number of successful outcomes among eight sampled generations.
        Prefix positions are measured in tokens from the beginning of the generated reasoning trajectory.
    }
    \label{tab:rtcv-additional-statistics}
    \centering
    \small
    \setlength{\tabcolsep}{5pt}
    \renewcommand{\arraystretch}{1.08}
    \begin{tabular}{@{}lcccc@{}}
        \toprule
        Subset &
        \shortstack{Source\\C/W} &
        \shortstack{Difficulty\\2/3/4/5/6} &
        \shortstack{Position median\\{}[\(p_{10}\), \(p_{90}\)]} &
        \shortstack{Candidates/prefix\\mean (median)} \\
        \midrule
        RCTV-Core
            & 252/248
            & 100/100/100/100/100
            & 2,259 [420, 6,224]
            & 5.43 (5.0) \\
        RCTV-Qwen8B
            & 236/264
            & 100/100/100/100/100
            & 2,960 [851, 6,641]
            & 6.30 (6.0) \\
        RCTV-Llama8B
            & 97/103
            & 40/40/40/40/40
            & 3,232 [800, 6,860]
            & 15.26 (15.0) \\
        RCTV-InterEnt
            & 252/248
            & 100/100/100/100/100
            & 1,894 [433, 5890]
            & 3.36 (3.0) \\
        \bottomrule
    \end{tabular}
\end{table*}

\begin{figure*}[t]
    \centering
    \includegraphics[width=0.92\textwidth]{
        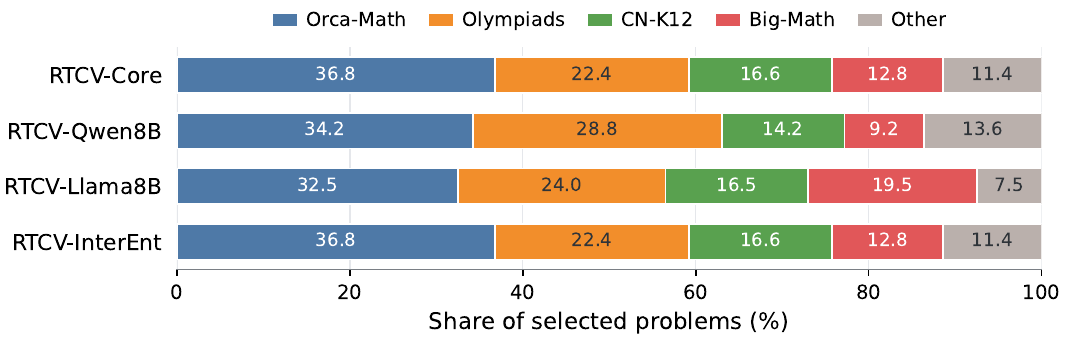
    }
    \caption{
        Original Big-Math source composition of the selected problems in
        each RCTV variant. The four most common sources are shown separately.
        Other comprises AOPS Forum, Math, OmniMath, HARP, GSM8K, and OpenMath.
        Percentages are normalized within each variant.
    }
    \label{fig:rtcv-source-composition}
\end{figure*}

\textbf{Problem-source composition.}
RCTV is constructed from problems drawn from Big-Math~\citep{albalak2025big}, which aggregates multiple mathematical reasoning datasets. Figure~\ref{fig:rtcv-source-composition} reports the original data sources of the problems retained after our difficulty-based selection. The selected problems span all major Big-Math sources, and no individual source accounts for more than 37\% of any RCTV variant. Problems are selected according to model difficulty, without imposing quotas on their original data sources. 

\label{app:rtcv-statistics}
\begin{figure*}[t]
\centering
\captionsetup[subfigure]{font=small}

\begin{subfigure}[t]{0.485\textwidth}
    \centering
    \includegraphics[width=\linewidth]{
        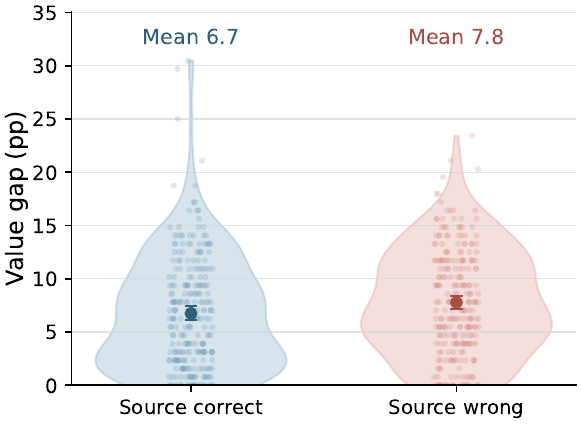
    }
    \caption{Value gaps by source outcome.}
    \label{fig:value-gap-source-outcome}
\end{subfigure}%
\hfill
\begin{subfigure}[t]{0.485\textwidth}
    \centering
    \includegraphics[width=\linewidth]{
        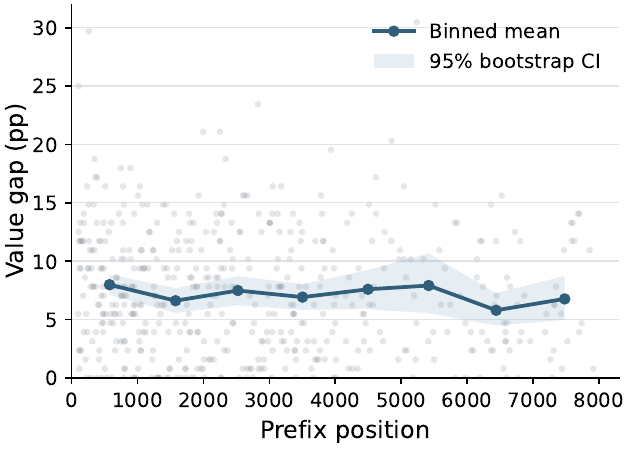
    }
    \caption{Value gaps across prefix positions.}
    \label{fig:value-gap-position}
\end{subfigure}

\caption{
    Additional analyses of within-prefix candidate-value variation.
    Left: Value-gap distributions grouped by whether the source trajectory
    produces a correct final answer.
    Right: Value gaps across prefix positions; the line shows binned means
    and the shaded region indicates 95\% bootstrap confidence intervals.
}
\label{fig:value-gap-additional}
\end{figure*}

\textbf{Value gaps across outcomes and positions.}
Figure~\ref{fig:value-gap-additional} examines how within-prefix candidate-value gaps vary with source-trajectory outcome and prefix position. Mean value gaps are modestly larger for incorrect than for correct source trajectories (7.8 versus 6.7 percentage points), indicating a slightly wider spread between higher- and lower-value candidates in the former group. However, the two distributions overlap substantially, so this difference is small. Thus, even prefixes from trajectories that ultimately fail can retain meaningful variation among plausible next-token choices, leaving room to select a candidate with better continuation prospects.
Across the first 8,000 generated positions, binned mean gaps remain approximately 6–8 percentage points and exhibit no clear monotonic trend. Overall, within the analyzed range, candidate-value variation does not consistently diminish as reasoning progresses.

\textbf{Candidate-value variation beyond sampling noise.}
We test whether the observed variation in candidate values could be explained by rollout-sampling noise alone. For each prefix, we compute the variance of the estimated values across its candidates and then average this variance across prefixes. Under the equal-value null, all candidates within a prefix share its pooled success probability, while the numbers of candidates and continuations are preserved. All 5,000 null simulations yielded a smaller average variance than the observed data (\(p=1/5001<2\times10^{-4}\)).
This result shows that the aggregate candidate variation cannot be explained by finite-rollout noise alone.

\textbf{Reliability of rollout-estimated values.}
We assess the reliability of rollout-estimated candidate values through the
posterior confidence in the empirical ordering of within-prefix candidate
pairs.
Let \(p(x,t)\) denote the underlying probability that a continuation sampled
after candidate \(t\) at prefix \(x\) produces a correct answer. Treating
continuation correctness as an independent Bernoulli outcome, for a
candidate with \(s\) successful continuations among \(R=128\) rollouts, we
place a Jeffreys prior on \(p(x,t)\),
\(p(x,t)\sim\operatorname{Beta}(\tfrac12,\tfrac12)\), yielding the posterior
\begin{equation}
p(x,t)\mid s
\sim
\operatorname{Beta}
\left(
s+\tfrac12,\,
R-s+\tfrac12
\right).
\label{eq:rollout-value-posterior}
\end{equation}
For each within-prefix candidate pair, we compute the posterior probability
that the candidate with the higher empirical value also has the higher
underlying success probability.

Table~\ref{tab:rollout-value-uncertainty} reports this directional
confidence across empirical-gap thresholds. At the main evaluation margin
\(m=0.08\), all 729 eligible pairs exceed \(0.90\) directional confidence,
with a median of \(0.971\). Thus, the candidate orderings used in the main
pairwise evaluation are well supported relative to finite rollout-sampling
uncertainty.

\begin{table*}[t]
\centering
\small
\caption{
Pairwise uncertainty in rollout-estimated candidate values.
Directional confidence is the posterior probability that the empirically
higher-valued candidate has the higher underlying success probability,
computed from all 128 rollouts using a Jeffreys prior.
The final column, labeled high-confidence (HC) prefixes, reports prefixes containing at least one pair with
directional confidence of at least \(0.90\).
}
\label{tab:rollout-value-uncertainty}
\begin{tabular}{lrccrc}
\toprule
Empirical gap
& \# Pairs
& \# Prefixes
& Median conf.
& Pairs \(\geq .90\)
& HC prefixes \\
\midrule
\(>0\)
& 5,515 & 460 & 0.794 & 26.2\% & 340 \\
\(\geq 0.04\)
& 2,043 & 328 & 0.915 & 56.2\% & 285 \\
\(\geq 0.08\)
& 729 & 199 & 0.971 & 100.0\% & 199 \\
\bottomrule
\end{tabular}
\end{table*}

\section{Model Specifications}
\label{app:model-specifications}
We use three reasoning models: Qwen3-1.7B, Qwen3-8B, and DeepSeek-R1-Distill-Llama-8B. DeepSeek-R1-Distill-Llama-8B is derived from Llama-3.1-8B-Base and fine-tuned on reasoning samples generated by DeepSeek-R1, providing a different model family from Qwen3.
Table~\ref{tab:model-details} summarizes their architectural specifications.
All model parameters remain frozen throughout probing.

\begin{table*}[t]
\caption{
    Architectural specifications of the reasoning models used in our
    experiments.
}
\label{tab:model-details}
\centering
\small
\setlength{\tabcolsep}{8pt}
\begin{tabular}{@{}lccc@{}}
\toprule
Model & Parameters & Number of layers & Hidden size \\
\midrule
Qwen3-1.7B & 1.7B & 28 & 2,048 \\
Qwen3-8B & 8.2B & 36 & 4,096 \\
DeepSeek-R1-Distill-Llama-8B & 8B & 32 & 4,096 \\
\bottomrule
\end{tabular}
\end{table*}

\begin{figure}[t]
\centering
  \includegraphics[width=0.8\columnwidth]{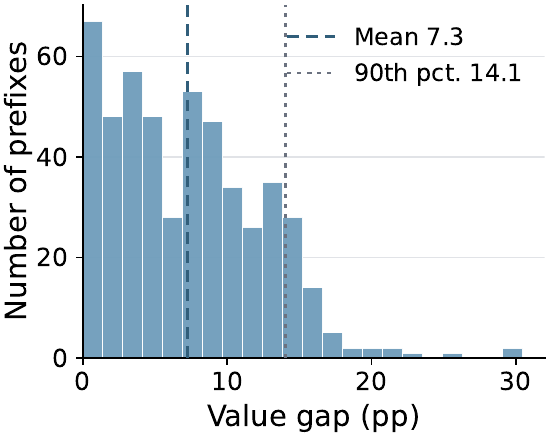}
  \caption{Distribution of value gaps in RCTV-Core.}
  \label{fig:value-gap-distribution}
\end{figure}

\section{Prefix-Token Interaction Representations}
\label{app:prefix-token-interactions}

We additionally construct a prefix-token interaction representation before \(t\) is processed by the model and examine whether relative long-horizon candidate value is linearly decodable from prefix-token interaction representations.
Given the output unembedding vector \(\mathbf{u}_t\) associated with token \(t\), and using \(\odot\) to denote element-wise multiplication, we define
\begin{equation}
\mathbf{z}_{\ell,r}^{\mathrm{int}}(x,t)
=
\mathbf{R}_{\ell,r}(x)_{|x|}
\odot
\mathbf{u}_t .
\label{eq:prefix-token-interaction}
\end{equation}
We apply the same five-fold probing procedure with ten problem-level fold seeds across layers and numbers of selected dimensions.

\begin{figure*}[t]
    \centering
    \includegraphics[width=\textwidth]
    {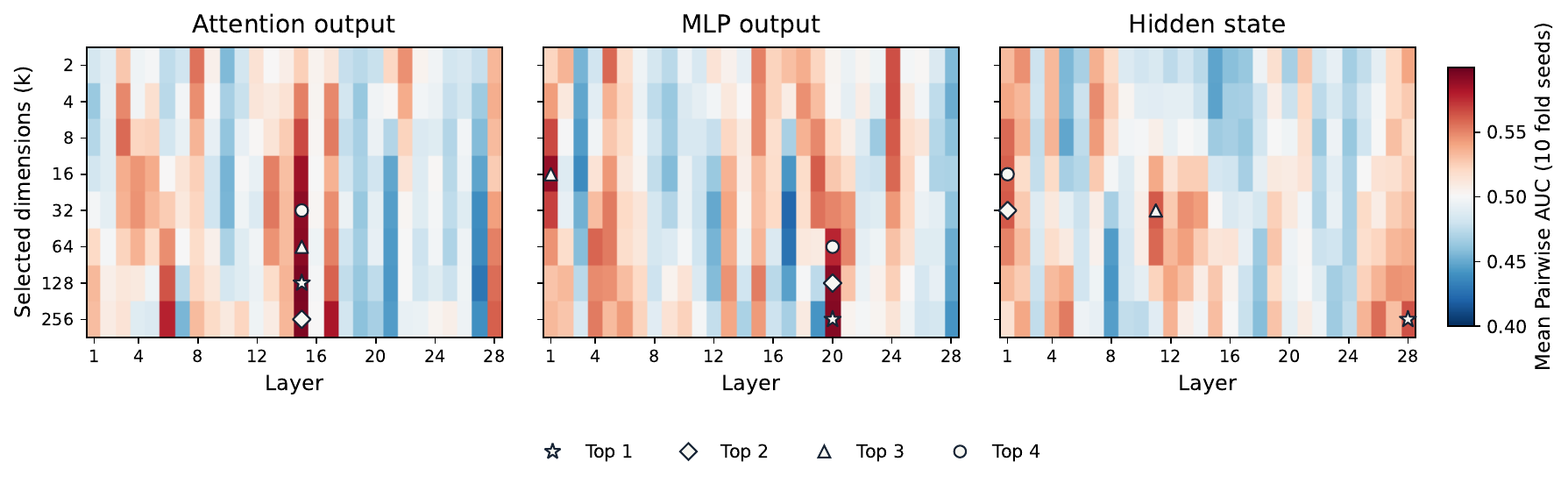}
    \caption{
    Candidate-value probing of prefix--token interaction
    representations across layers and selected dimensionalities.
    Each cell reports the mean pairwise AUC over ten problem-level
    fold seeds for attention outputs, MLP outputs, and hidden states.
    Within each representation family, \(\star\), \(\diamond\),
    \(\triangle\), and \(\circ\) mark the first- through fourth-highest
    observed mean pairwise AUC, respectively.
    }
    \label{fig:prefix-token-heatmaps}
\end{figure*}

\begin{table}[t]
\centering
\small
\setlength{\tabcolsep}{2.5pt}
\begin{tabular}{lrrrr}
\toprule
Repr & \(k\) & Pairwise \(\uparrow\) & Tier \(\uparrow\)
& Regret@1 \(\downarrow\) \\
\midrule
Attn (\(\mathrm{A}_{15}\)) & 128
& 59.4 {\scriptsize$\pm$ 3.2}
& 57.5 {\scriptsize$\pm$ 2.2}
& 0.0328 {\scriptsize$\pm$ 0.0012} \\
MLP (\(\mathrm{M}_{20}\)) & 256
& 59.2 {\scriptsize$\pm$ 2.6}
& 57.8 {\scriptsize$\pm$ 2.5}
& 0.0322 {\scriptsize$\pm$ 0.0008} \\
Hidden (\(\mathrm{H}_{28}\)) & 256
& 56.4 {\scriptsize$\pm$ 2.0}
& 54.3 {\scriptsize$\pm$ 2.0}
& 0.0345 {\scriptsize$\pm$ 0.0014} \\
\bottomrule
\end{tabular}
\caption{
Candidate-value readout results for prefix-token interaction
representations. We report the mean and standard deviation over ten problem-level fold seeds. For each representation family, we report the strongest observed configuration according to mean pairwise AUC.
}
\label{tab:prefix-token-results}
\end{table}

Figure~\ref{fig:prefix-token-heatmaps} shows higher pairwise AUC near layer 15 for attention outputs and layer 20 for MLP outputs.
For hidden states, the interaction yields above-chance readout performance in late layers, with the highest mean pairwise AUC
at layer 28 and \(k=256\).
Table~\ref{tab:prefix-token-results} reports the strongest observed configuration in each family.

\section{Sparse Readout Procedure}
\label{app:sparse-readout}

\textbf{Feature selection.}
For each fold \(f\), we compute the feature-selection score for dimension
\(d\) using only the training portion:
\begin{equation}
s_{f,d}
=
\left|
\operatorname{Corr}_{(x,t)\in
\mathcal{D}_{\mathrm{train}}^{(f)}}
\left(
z_d(x,t),\Delta \hat v(x,t)
\right)
\right|.
\label{eq:dimension-selection-score}
\end{equation}
Here, \(\operatorname{Corr}\) denotes the sample Pearson correlation over all candidate instances in the corresponding training fold.
The selected dimension set is
\begin{equation}
\mathcal{I}_{f,k}
\in
\underset{
    \substack{
        \mathcal{I}\subseteq\{1,\ldots,D\}\\
        |\mathcal{I}|=k
    }
}{
    \operatorname{arg\,max}
}
\sum_{d\in\mathcal{I}} s_{f,d},
\label{eq:topk-dimension-selection}
\end{equation}
where \(D\) is the dimensionality of the representation. For multi-site
readouts, the constituent representations are concatenated before selection,
and dimensions are selected jointly from the resulting feature space.

\textbf{Optimization procedure.}
After feature selection, we standardize the selected dimensions using their
training-fold means and standard deviations. 
For a fixed layer \(\ell\) and representation type \(r\),
let \(\mathbf z^{\mathrm{std}}_{f,\mathcal I_{f,k}}(x,t)\)
denote \(\mathbf z_{\ell,r}(x,t)\) restricted to the selected
dimensions and standardized using the training-fold mean and standard deviation of each dimension.
The same transformation is applied to held-out candidates. Let \(\mathbf y_f\) collect the training-fold targets \(\Delta\hat v(x,t)\), and let \(\mathbf Z_{f,k}\) denote
the corresponding design matrix of standardized features restricted to the selected dimensions. \(\mathbf 1\) denotes the all-ones vector. We fit the ridge probe by minimizing the regularized squared-error objective in Eq.~\ref{eq:sparse-probe-objective}. We fix the ridge regularization coefficient to \(\lambda=1\) in all experiments. The training-fold standardization and fitted readout are then applied to the held-out fold. Repeating this procedure across the five folds yields one out-of-fold prediction for every candidate, from which all evaluation metrics are computed.
\begin{equation}
\begin{gathered}
s_{f,k}(x,t;\mathbf{w},b)
=
\mathbf{w}^{\top}
\mathbf{z}^{\mathrm{std}}_{f,\mathcal{I}_{f,k}}(x,t)+b
\\[5pt]
\mathcal{L}_{f,k}(\mathbf{w},b)
=
\|\mathbf{y}_f-\mathbf{Z}_{f,k}\mathbf{w}-b\mathbf{1}\|_2^2
+\lambda\|\mathbf{w}\|_2^2
\\[5pt]
(\widehat{\mathbf{w}}_{f,k},\widehat b_{f,k})
=
\underset{\mathbf{w},b}{\operatorname{arg\,min}}\,
\mathcal{L}_{f,k}(\mathbf{w},b).
\end{gathered}
\label{eq:sparse-probe-objective}
\end{equation}

\section{Threshold Statistics and Sensitivity}
\label{app:threshold-sensitivity}

\textbf{Threshold-induced sample sizes.}
Table~\ref{tab:threshold-statistics} reports how the pairwise margin \(m\) and tier threshold \(\epsilon\) affect the evaluation sample on RCTV-Core.
At the main pairwise margin \(m=0.08\), the evaluation retains 729 pairs from 199 prefixes. 
At the main tier threshold \(\epsilon=0.04\), it retains 513 candidates from 249 prefixes, with similar numbers of low- and high-value candidates (251 and 262, respectively). Thus, the main thresholds exclude near-ties while retaining substantial and approximately balanced evaluation sets. 
All confidence intervals in main experiments are computed by resampling at the problem level.

\begin{table*}[t]

\centering
\small

\begin{minipage}[t]{0.47\textwidth}
\centering
\textit{Pairwise-margin statistics}\\[3pt]
\begin{tabular}{@{}rrrr@{}}
\toprule
\(m\) & Pairs & Coverage (\%) & Prefixes \\
\midrule
0.00 & 5,515 & 100.0 & 460 \\
0.02 & 3,724 & 67.5 & 413 \\
0.04 & 2,043 & 37.0 & 328 \\
0.06 & 1,389 & 25.2 & 280 \\
\(\mathbf{0.08}\) & \textbf{729} & \textbf{13.2} & \textbf{199} \\
0.10 & 460 & 8.3 & 152 \\
\bottomrule
\end{tabular}
\end{minipage}
\hfill
\begin{minipage}[t]{0.49\textwidth}
\centering
\textit{Value-tier statistics}\\[3pt]
\begin{tabular}{@{}rrrrr@{}}
\toprule
\(\epsilon\) & Low & High & Coverage (\%) & Prefixes \\
\midrule
0.00 & 1,219 & 1,255 & 91.1 & 460 \\
0.01 & 830 & 842 & 61.6 & 419 \\
0.02 & 559 & 589 & 42.3 & 365 \\
0.03 & 389 & 374 & 28.1 & 298 \\
\(\mathbf{0.04}\) & \textbf{251} & \textbf{262} &
    \textbf{18.9} & \textbf{249} \\
0.05 & 164 & 172 & 12.4 & 185 \\
\bottomrule
\end{tabular}
\end{minipage}
\caption{
    Evaluation sample sizes under different thresholds on RCTV-Core.
    Exact value ties are excluded from the pairwise evaluation at every
    \(m\), and pair coverage is measured relative to the 5,515 non-tied
    within-prefix pairs.
    Tier coverage is the fraction of the 2,715 candidates satisfying
    \(\lvert\Delta\hat v(x,t)\rvert>\epsilon\).
    The prefix columns report the numbers of the 500 prefixes contributing at
    least one retained pair or tier candidate.
    Bold rows indicate the thresholds used in the main analysis.
}
\label{tab:threshold-statistics}
\end{table*}

\begin{figure*}[t]
    \centering
    \includegraphics[width=\textwidth]
    {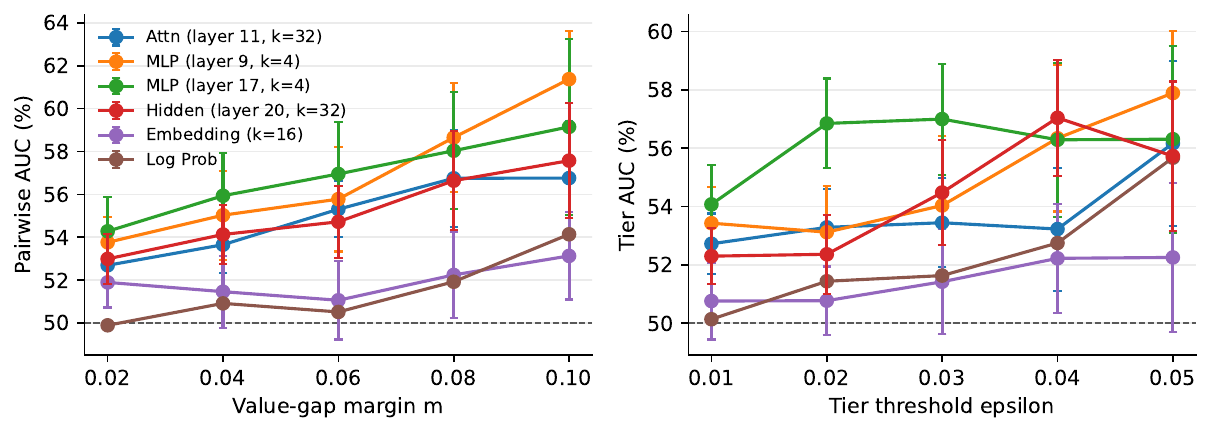}
    \caption{Sensitivity of candidate-value readout performance
    to the value-gap margin \(m\) (left) and tier threshold
    \(\epsilon\) (right) on RCTV-Core.
    Points and error bars indicate the mean and standard deviation
    across ten problem-level fold seeds.
    Log Prob uses untrained next-token log probabilities and
    therefore does not vary across fold seeds.
    Dashed lines indicate chance-level AUC.}
    \label{fig:margin-epsilon-sensitivity}
\end{figure*}

\textbf{Threshold sensitivity.}
Figure~\ref{fig:margin-epsilon-sensitivity} shows that candidate-position readouts outperform both baselines in pairwise AUC across all evaluated margins.
They also achieve higher mean tier AUC than both baselines across all evaluated tier thresholds.
These comparisons indicate that the observed advantages are not limited to the thresholds used in the main analysis.

\section{Additional Heatmaps of Candidate-Value Readout}
\label{app:readout-heatmaps}

\begin{figure*}[t]
    \centering
    \includegraphics[width=\textwidth]
    {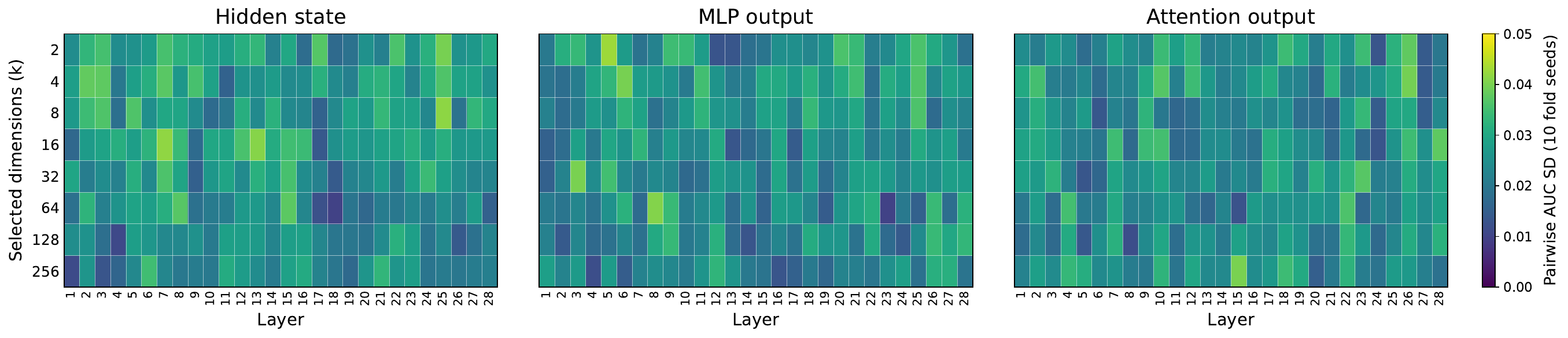}
    \caption{Standard deviation of pairwise AUC across ten
    problem-level fold seeds for candidate-value readouts.
    Each cell corresponds to a layer and selected dimensionality.}
    \label{fig:readout-sd}
\end{figure*}

\begin{figure*}[t]
    \centering
    \includegraphics[width=\textwidth]
    {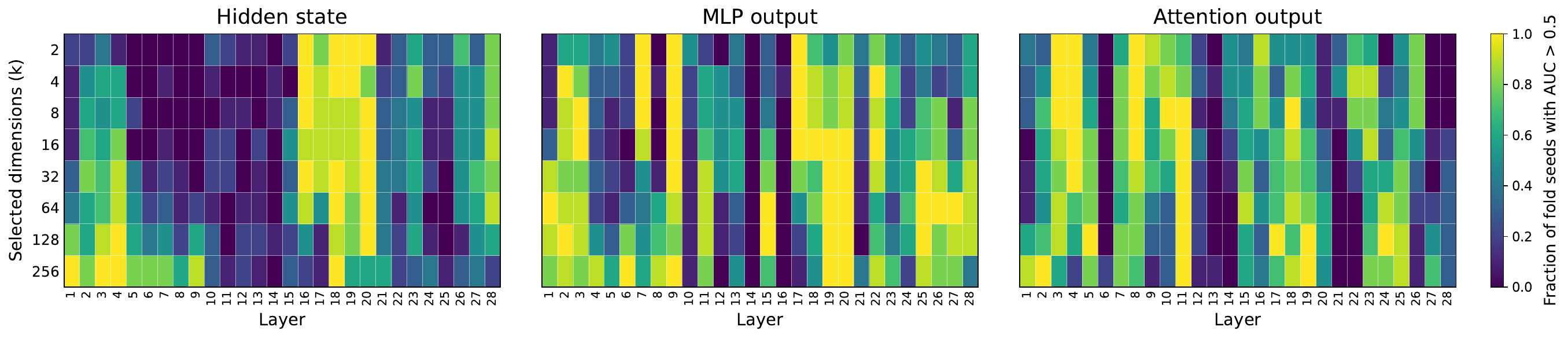}
    \caption{Fraction of the ten problem-level fold seeds
    with pairwise AUC above \(0.5\) for candidate-value
    readouts.}
    \label{fig:readout-above-chance}
\end{figure*}

We supplement the mean pairwise AUC heatmaps with two summaries across the same ten problem-level fold seeds.
Figure~\ref{fig:readout-sd} reports the standard deviation of pairwise AUC for each layer--dimensionality configuration.
Figure~\ref{fig:readout-above-chance} reports the fraction of fold seeds with pairwise AUC above \(0.5\).
Together, these summaries characterize variation in readout performance and the consistency of above-chance results across fold assignments.

\section{Candidate-value Readout Across Representation Types in Additional Models}
\label{app:cross-model-representations}

\begin{figure*}[t]
    \centering
    \includegraphics[width=\textwidth]
    {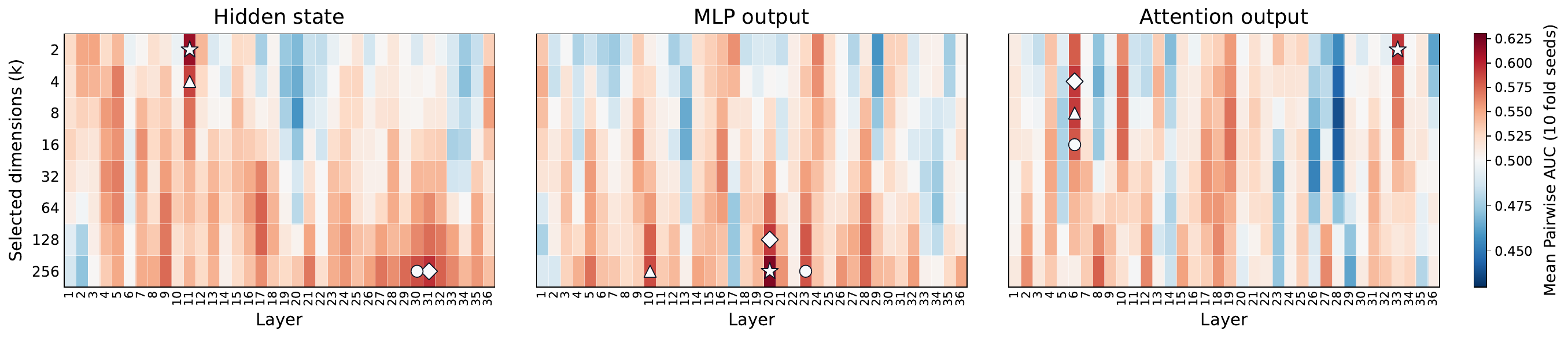}
    \caption{Candidate-value readout from hidden states,
    MLP outputs, and attention outputs in Qwen3-8B.
    Each cell reports mean pairwise AUC across ten
    problem-level fold seeds.
    Within each representation family, \(\star\), \(\diamond\),
    \(\triangle\), and \(\circ\) mark the first- through
    fourth-highest observed mean pairwise AUC, respectively.}
    \label{fig:qwen8b-all-representations}
\end{figure*}

\begin{figure*}[t]
    \centering
    \includegraphics[width=\textwidth]
    {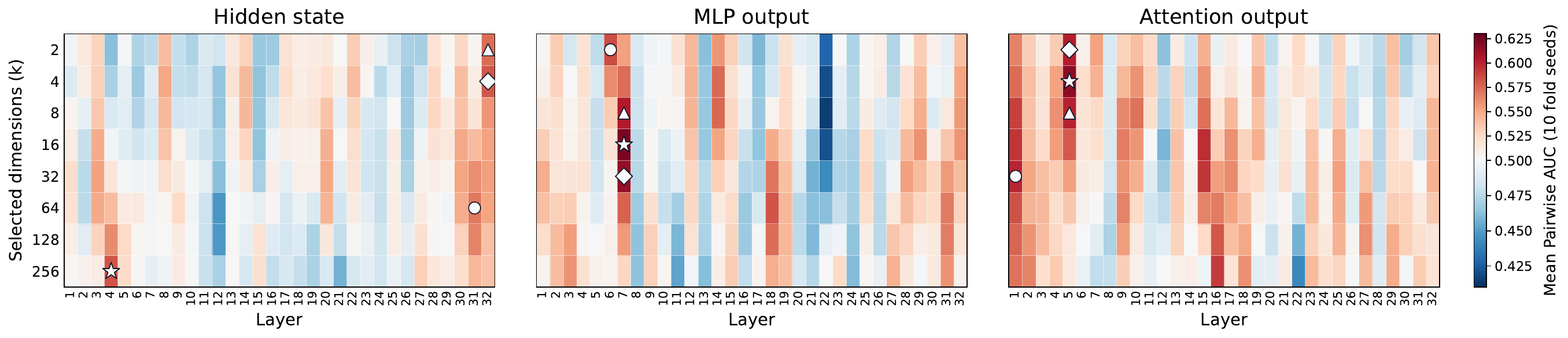}
    \caption{Candidate-value readout from hidden states,
    MLP outputs, and attention outputs in DS-Llama-3.1-8B.
    Each cell reports mean pairwise AUC across ten
    problem-level fold seeds.
    Within each representation family, \(\star\), \(\diamond\),
    \(\triangle\), and \(\circ\) mark the first- through
    fourth-highest observed mean pairwise AUC, respectively.}
    \label{fig:llama8b-all-representations}
\end{figure*}

We extend the cross-model analysis beyond MLP outputs
to hidden states and attention outputs, using the same
probing procedure and ten problem-level fold seeds.
Figures~\ref{fig:qwen8b-all-representations}
and~\ref{fig:llama8b-all-representations}
report mean pairwise AUC across layers and selected
dimensionalities.
Both Qwen3-8B and DS-Llama-3.1-8B exhibit above-chance mean pairwise AUC at multiple layers across all three representation types.
Tables~\ref{tab:qwen8b-candidate-value-readout} and~\ref{tab:llama8b-candidate-value-readout} summarize the strongest observed configuration in each representation family and the baselines.
In both models, all three reported readouts achieve higher pairwise and tier AUC and lower Regret@1 than all evaluated baselines.
Together, these results provide evidence that candidate value is linearly decodable from candidate-position representations in multiple reasoning models.
They further illustrate that readout patterns vary across models and representation types.

\begin{table}[!htbp]
\small
\setlength{\tabcolsep}{2.8pt}
\begin{tabular}{lrccc}
\toprule
Repr
& $k$
& Pairwise $\uparrow$
& Tier $\uparrow$
& Regret@1 $\downarrow$\\
\midrule

\rowcolor{black!7}
\multicolumn{5}{l}{\textit{Baselines}} \\[2pt]

Uniform
& -
& 50.0
& 50.0
& 0.0386 \\

Log prob
& 1
& 48.2
& 48.5
& 0.0412 \\

Prob rank
& 1
& 52.5
& 49.5
& 0.0412 \\

Embedding
& 64
& 55.1 {\scriptsize$\pm$ 1.8}
& 53.0 {\scriptsize$\pm$ 1.3}
& 0.0393 {\scriptsize$\pm$ 0.0010} \\

\midrule
\rowcolor{black!7}
\multicolumn{5}{l}{\textit{Candidate-value readouts}} \\[2pt]

Hidden ($\mathrm{H}_{11}$)
& 2
& 61.2 {\scriptsize$\pm$ 2.0}
& 57.1 {\scriptsize$\pm$ 1.8}
& 0.0363 {\scriptsize$\pm$ 0.0011} \\

MLP ($\mathrm{M}_{20}$)
& 256
& \textbf{62.0} {\scriptsize$\pm$ 2.6}
& 57.2 {\scriptsize$\pm$ 2.1}
& \textbf{0.0356} {\scriptsize$\pm$ 0.0012} \\

Attn ($\mathrm{A}_{33}$)
& 2
& 59.3 {\scriptsize$\pm$ 2.0}
& \textbf{58.1} {\scriptsize$\pm$ 2.0}
& 0.0360 {\scriptsize$\pm$ 0.0007} \\

\bottomrule
\end{tabular}
\caption{
Candidate-value readout results and baselines for Qwen3-8B.
For fitted readouts, we report the mean and standard deviation
over ten problem-level fold seeds; parameter-free baselines
are reported as single values.
For each internal-representation family and the embedding
baseline, we report the strongest observed configuration
according to mean pairwise AUC.
}
\label{tab:qwen8b-candidate-value-readout}
\end{table}

\begin{table}[!htbp]
\small
\setlength{\tabcolsep}{2.8pt}
\begin{tabular}{lrccc}
\toprule
Repr
& $k$
& Pairwise $\uparrow$
& Tier $\uparrow$
& Regret@1 $\downarrow$\\
\midrule

\rowcolor{black!7}
\multicolumn{5}{l}{\textit{Baselines}} \\[2pt]

Uniform
& -
& 50.0
& 50.0
& 0.0490 \\

Log prob
& 1
& 50.9
& 50.1
& 0.0509 \\

Prob rank
& 1
& 50.7
& 50.5
& 0.0509 \\

Embedding
& 128
& 52.6 {\scriptsize$\pm$ 1.7}
& 53.2 {\scriptsize$\pm$ 1.5}
& 0.0504 {\scriptsize$\pm$ 0.0013} \\

\midrule
\rowcolor{black!7}
\multicolumn{5}{l}{\textit{Candidate-value readouts}} \\[2pt]

Hidden ($\mathrm{H}_{4}$)
& 256
& 58.1 {\scriptsize$\pm$ 1.9}
& 57.6 {\scriptsize$\pm$ 2.4}
& 0.0462 {\scriptsize$\pm$ 0.0016} \\

MLP ($\mathrm{M}_{7}$)
& 16
& \textbf{62.3} {\scriptsize$\pm$ 1.8}
& 59.0 {\scriptsize$\pm$ 2.0}
& 0.0459 {\scriptsize$\pm$ 0.0025} \\

Attn ($\mathrm{A}_{5}$)
& 4
& 61.6 {\scriptsize$\pm$ 2.5}
& \textbf{59.5} {\scriptsize$\pm$ 1.7}
& \textbf{0.0421} {\scriptsize$\pm$ 0.0018} \\

\bottomrule
\end{tabular}
\caption{
Candidate-value readout results and baselines for DS-Llama-3.1-8B.
For fitted readouts, we report the mean and standard deviation
over ten problem-level fold seeds; parameter-free baselines
are reported as single values.
For each internal-representation family and the embedding
baseline, we report the strongest observed configuration
according to mean pairwise AUC.
}
\label{tab:llama8b-candidate-value-readout}
\end{table}

\section{Candidate-Value Readout in the Intermediate-Entropy Setting}
\label{app:entropy-analysis}

We examine whether dimensions identified on RCTV-Core also support candidate-value readout on RCTV-InterEnt.
We fix the dimension sets to \(\{1834,312\}\) and \(\{1834,312,2021,822\}\) at \(\mathrm{M}_{17}\), and \(\{432,950\}\) and \(\{432,950,256,1274\}\) at
\(\mathrm{M}_{9}\).
For each set, we evaluate ridge, signed-sum, and correlation-weighted readouts using five-fold cross-validation with ten problem-level fold seeds.
Dimensions remain fixed, while decoder weights are estimated using each InterEnt training fold.

Table~\ref{tab:interent-fixed-core-readouts} reports the results alongside parameter-free baselines.
The two-dimensional configuration at \(\mathrm{M}_{17}\) achieves higher mean pairwise and tier AUC and lower Regret@1 than all three baselines with every decoder.
At \(\mathrm{M}_{9}\), both dimension sets achieve higher mean tier AUC and lower Regret@1 than all three baselines with every decoder, although pairwise AUC is weaker.
These results provide evidence that Core-selected dimensions also support candidate-value readout in the intermediate-entropy setting, while readout performance varies across layers, dimension sets, and metrics.
Log-probability and rank baselines also achieve above-chance AUC in InterEnt, indicating that local token preferences provide some discrimination of long-horizon candidate value in this intermediate-entropy subset.

\begin{table*}[!htbp]
\centering
\small
\setlength{\tabcolsep}{6pt}
\begin{tabular}{llrccc}
\toprule
Repr
& Decoder
& $k$
& Pairwise $\uparrow$
& Tier $\uparrow$
& Regret@1 $\downarrow$ \\
\midrule
\rowcolor{black!7}
\multicolumn{6}{l}{\textit{Baselines}} \\[2pt]

Uniform & - & -
& 50.0 & 50.0 & 0.0275 \\

Log prob & - & 1
& 54.9 & 54.9 & 0.0281 \\

Prob rank & - & 1
& 55.2 & 53.3 & 0.0281 \\

\midrule
\rowcolor{black!7}
\multicolumn{6}{l}{\textit{Fixed Core-selected dimensions at $\mathrm{M}_{17}$}} \\[2pt]

MLP ($\mathrm{M}_{17}$) & Ridge & 2
& 57.3 {\scriptsize$\pm$ 2.9}
& 56.1 {\scriptsize$\pm$ 2.3}
& 0.0241 {\scriptsize$\pm$ 0.0006} \\

& Signed-sum & 2
& 60.5 {\scriptsize$\pm$ 0.1}
& 58.4 {\scriptsize$\pm$ 0.1}
& 0.0243 {\scriptsize$\pm$ 0.0000} \\

& Correlation-weighted & 2
& 58.7 {\scriptsize$\pm$ 2.4}
& 57.2 {\scriptsize$\pm$ 1.5}
& 0.0241 {\scriptsize$\pm$ 0.0005} \\[2pt]

& Ridge & 4
& 52.2 {\scriptsize$\pm$ 2.6}
& 52.2 {\scriptsize$\pm$ 2.1}
& 0.0256 {\scriptsize$\pm$ 0.0006} \\

& Signed-sum & 4
& 51.0 {\scriptsize$\pm$ 2.2}
& 49.4 {\scriptsize$\pm$ 1.4}
& 0.0266 {\scriptsize$\pm$ 0.0009} \\

& Correlation-weighted & 4
& 54.5 {\scriptsize$\pm$ 2.3}
& 53.7 {\scriptsize$\pm$ 2.0}
& 0.0249 {\scriptsize$\pm$ 0.0007} \\

\midrule
\rowcolor{black!7}
\multicolumn{6}{l}{\textit{Fixed Core-selected dimensions at $\mathrm{M}_{9}$}} \\[2pt]

MLP ($\mathrm{M}_{9}$) & Ridge & 2
& 51.3 {\scriptsize$\pm$ 1.6}
& 57.7 {\scriptsize$\pm$ 1.0}
& 0.0259 {\scriptsize$\pm$ 0.0005} \\

& Signed-sum & 2
& 53.2 {\scriptsize$\pm$ 0.8}
& 57.9 {\scriptsize$\pm$ 0.4}
& 0.0252 {\scriptsize$\pm$ 0.0003} \\

& Correlation-weighted & 2
& 52.3 {\scriptsize$\pm$ 1.2}
& 58.3 {\scriptsize$\pm$ 1.0}
& 0.0259 {\scriptsize$\pm$ 0.0004} \\[2pt]

& Ridge & 4
& 47.6 {\scriptsize$\pm$ 3.1}
& 55.3 {\scriptsize$\pm$ 3.4}
& 0.0273 {\scriptsize$\pm$ 0.0011} \\

& Signed-sum & 4
& 51.7 {\scriptsize$\pm$ 4.0}
& 57.4 {\scriptsize$\pm$ 3.5}
& 0.0258 {\scriptsize$\pm$ 0.0008} \\

& Correlation-weighted & 4
& 48.1 {\scriptsize$\pm$ 2.8}
& 55.9 {\scriptsize$\pm$ 3.0}
& 0.0270 {\scriptsize$\pm$ 0.0010} \\

\bottomrule
\end{tabular}
\caption{
Readout results with fixed Core-selected dimensions on RCTV-InterEnt.
The dimension sets are \(\{1834,312\}\) and
\(\{1834,312,2021,822\}\) at \(\mathrm{M}_{17}\), and
\(\{432,950\}\) and \(\{432,950,256,1274\}\) at \(\mathrm{M}_{9}\).
Decoder weights are estimated
using each InterEnt training fold.
We report mean and standard deviation over ten problem-level fold seeds; parameter-free baselines are reported as single values.
}
\label{tab:interent-fixed-core-readouts}
\end{table*}

\section{Online Token Interventions: Additional Details and Results}
\label{app:online-interventions}
We next move from offline probing to online generation and test whether probe estimates of relative candidate value can guide token interventions during generation. In the offline analysis, probes only rank candidates associated with fixed RCTV prefixes. In the online setting, probe predictions determine which candidate is appended during generation, thereby changing the context from which subsequent tokens are generated. This evaluates whether the decoded value signal remains useful as the reasoning trajectory evolves.


\subsection{Intervention Setup}
\label{app:online-interventions_setup}
\textbf{Readout configurations and fitting.}
We fix all intervention configurations before observing any online intervention outcomes.
To fit the final intervention readouts, we split the 500 RCTV-Core problems at the problem level.
We train the readouts on 400 RCTV-Core problems and evaluate the intervention strategies on newly generated trajectories for the remaining 100.
Dimension selection, feature standardization, and ridge coefficients are all estimated using only these 400 problems.
The intervened readouts are fixed before intervention trajectories are generated for the 100 evaluation problems.

\textbf{Probe-guided strategy.}
We construct the candidate-value readout using the same procedure as in the offline analysis.
We use \(\mathrm{M}_{17}\) with \(k=4\), one of the high-performing sparse configurations identified in that analysis.
At each intervention opportunity, the strategy selects the candidate with the highest predicted relative value.

\textbf{Other strategies.}
We consider top-probability selection, which chooses the candidate with the highest next-token probability, and
uniform-random selection, which samples uniformly from the plausible candidate set.
The no-intervention baseline follows standard next-token sampling throughout the trajectory.

\textbf{Intervention protocol.}
For a controlled comparison, all strategies first follow standard sampling to produce the same initial prefix. They then apply their respective candidate-selection rules at high-entropy positions, with sampling randomness shared throughout each trajectory. 
Specifically, we use the first 4,000 generated tokens as the common initial prefix and extend each trajectory to 8,000 tokens under the intervention protocol and strategies. Within this window, intervention opportunities and plausible candidate sets follow the same entropy threshold and probability criteria as in the main offline analysis. 
Interventions are applied online: after a selected token is appended, generation continues from the resulting prefix.
We denote by \(B\) the maximum number of interventions per trajectory and evaluate intervention budgets \(B\in\{8,16,32,64\}\).
Generation follows standard sampling at all other positions. 

\textbf{Continuation sampling and verification.}
For each of the 100 evaluation problems, we generate two independent
trajectory runs under each strategy, with sampling randomness matched across strategies within each run.
Each intervention trajectory produces an 8,000-token prefix. We estimate its long-horizon value using 128 independently sampled continuations and determine correctness using the same verification procedure as in RCTV construction. 
The primary results aggregate all trajectories across the 100 evaluation problems. Supplementary analyses examine consistency across independent trajectory generations and across problem-difficulty groups.

\textbf{Outcome metrics.}
We report two outcome metrics. Mean continuation accuracy is the fraction of the 128 continuations that produce a correct final answer, averaged across prefixes and problems. It estimates the success probability of a single continuation from an intervention prefix. Prefix-conditioned Pass@\(k\) measures the probability that at least one of \(k\) continuations from the same intervention prefix yields a correct final answer. We compute the unbiased Pass@\(k\) estimator for each prefix, average over the prefixes. We adopt the pass@\(k\) formulation of \citet{chen2021evaluating} for consistency with prior work. We report Pass@\(k\) for \(k\in\{4,8,16,32\}\). 

\subsection{Additional Intervention Results}
\label{app:additional-intervention-results}

\textbf{Evaluation on problems excluded from offline analysis.}
We additionally compare probe-guided intervention with standard sampling on the 61 evaluation problems outside the offline pairwise-evaluation subset.
These problems contribute no eligible candidate pairs at the main value-gap margin \(m=0.08\), and therefore do not contribute to the pairwise AUC used to select the readout configuration.
This analysis examines whether online gains extend beyond the problems contributing to that selection metric. Table~\ref{tab:intervention-pairwise-excluded} reports
mean continuation accuracy and paired differences.
Probe-guided intervention achieves higher mean accuracy at all four budgets, with gains ranging from 2.16 to 5.71 percentage points.
The paired 95\% bootstrap confidence intervals exclude zero at \(B=32\) and \(B=64\), and include zero at \(B=8\) and \(B=16\).

\begin{table}[t]
\centering
\small
\setlength{\tabcolsep}{3pt}
\begin{tabular}{rlcc}
\toprule
$B$ & Strategy & Avg@128 (\%) & $\Delta$ [95\% CI] \\
\midrule
0  & Standard     & 54.93 & --- \\
8  & Probe-guided & 57.10 & +2.16 [-1.76, +6.28] \\
16 & Probe-guided & 57.42 & +2.49 [-2.04, +7.79] \\
32 & Probe-guided & 60.64 & +5.71 [+1.20, +10.58] \\
64 & Probe-guided & 59.63 & +4.70 [+0.65, +9.54] \\
\bottomrule
\end{tabular}
\caption{
Online intervention results on the 61 problems outside the
offline pairwise-evaluation subset.
Continuation accuracy is reported in percent, and differences
relative to standard sampling are reported in percentage points
with paired 95\% bootstrap confidence intervals.
Standard sampling involves no interventions (\(B=0\)).
}
\label{tab:intervention-pairwise-excluded}
\end{table}

\textbf{Results across trajectory seeds.}
We report online intervention results separately for the two trajectory seeds.
For each problem and trajectory seed, strategies share the same random draws for next-token sampling at corresponding token positions.
Table~\ref{tab:intervention-trajectory-seeds} compares probe-guided interventions with standard sampling.
Although standard-sampling accuracy differs between the two seeds, probe-guided interventions achieve higher continuation accuracy than the corresponding standard-sampling baseline at every evaluated budget in both seeds.
The magnitude of the gains varies across seeds and budgets.

\begin{table}[t]
\centering
\small
\setlength{\tabcolsep}{5pt}
\begin{tabular}{rlcc}
\toprule
$B$ & Strategy & Trajectory 0 & Trajectory 1 \\
\midrule
0  & Standard     & 46.98 & 53.44 \\
\midrule
8  & Probe-guided & 47.49 & 54.12 \\
16 & Probe-guided & 48.34 & 56.23 \\
32 & Probe-guided & 50.87 & 56.87 \\
64 & Probe-guided & 52.35 & 56.67 \\
\bottomrule
\end{tabular}
\caption{
Continuation accuracy (\%) reported separately for the two
trajectory seeds, estimated using 128 continuations per prefix.
}
\label{tab:intervention-trajectory-seeds}
\end{table}

\textbf{Prefix-conditioned Pass@\(k\).}
We further evaluate intervention outcomes using prefix-conditioned Pass@\(k\), which measures the probability that at least one of \(k\) continuations from the resulting
prefix succeeds.
Table~\ref{tab:online-prefix-pass-at-k} reports estimates for \(k\in\{4,8,16,32\}\), computed from 128 continuations per prefix and averaged across evaluation problems.
Probe-guided interventions achieve higher Pass@\(k\) than standard sampling at all evaluated budgets and values of \(k\).
They also achieve higher point estimates than both top-probability and uniform-random selection at \(B=32\) and \(B=64\).

\begin{table}[!htbp]
\centering
\small
\setlength{\tabcolsep}{3pt}
\begin{tabular}{rlrrrr}
\toprule
$B$ & Strategy & Pass@4 & Pass@8 & Pass@16 & Pass@32 \\
\midrule
0 & Standard & 57.14 & 59.54 & 61.35 & 62.82 \\
\midrule
8 & Top-prob     & 59.37 & 62.16 & 64.31 & 65.91 \\
  & Uniform      & 58.11 & 60.86 & 62.92 & 64.51 \\
  & Probe-guided & 57.78 & 60.64 & 62.61 & 63.89 \\
\midrule
16 & Top-prob     & 57.01 & 59.72 & 61.70 & 63.06 \\
   & Uniform      & 58.77 & 61.01 & 62.62 & 63.93 \\
   & Probe-guided & 58.50 & 61.19 & 63.64 & 65.61 \\
\midrule
32 & Top-prob     & 58.13 & 60.40 & 61.82 & 62.78 \\
   & Uniform      & 56.96 & 59.13 & 60.83 & 61.97 \\
   & Probe-guided & 60.06 & 62.33 & 64.23 & 65.92 \\
\midrule
64 & Top-prob     & 57.56 & 60.04 & 61.63 & 62.71 \\
   & Uniform      & 59.21 & 61.73 & 63.52 & 64.95 \\
   & Probe-guided & 61.18 & 63.18 & 64.39 & 65.32 \\
\bottomrule
\end{tabular}
\caption{
Prefix-conditioned Pass@\(k\) (\%) across intervention budgets.
Standard sampling involves no interventions.
}
\label{tab:online-prefix-pass-at-k}
\end{table}

\textbf{Paired comparisons of continuation accuracy.} Table~\ref{tab:online_mean_continuation_accuracy} reports continuation accuracy and paired comparisons among probe-guided interventions, standard sampling,
uniform-random selection, and top-probability selection.
Probe-guided continuation accuracy increases with the intervention budget.
At \(B=64\), probe-guided interventions improve accuracy by 4.30 percentage points over standard sampling and 3.96 percentage points over top-probability selection,
with both paired 95\% bootstrap confidence intervals excluding zero.
The estimated gain over uniform-random selection is 2.67 percentage points, with an interval that includes zero.

\begin{table*}[!htbp]
\centering
\footnotesize
\setlength{\tabcolsep}{3.5pt}
\begin{tabular}{@{}c l c c c c@{}}
\toprule
\(\boldsymbol{B}\) &
\textbf{Policy} &
\textbf{Avg@128} &
\(\boldsymbol{\Delta}\) \textbf{vs.\ Standard [95\% CI]} &
\(\boldsymbol{\Delta}\) \textbf{vs.\ Uniform [95\% CI]} &
\(\boldsymbol{\Delta}\) \textbf{vs.\ Top-prob [95\% CI]} \\
\midrule

0
& Standard (no intervention)
& 50.21
& ---
& ---
& --- \\

\midrule

\multirow{3}{*}{8}
& Uniform
& 51.63
& +1.42 [-2.36, +5.10]
& ---
& -0.61 [-3.74, +2.43] \\
& Top-prob
& 52.23
& +2.03 [-1.54, +5.70]
& +0.61 [-2.43, +3.74]
& --- \\
& Probe-guided
& 50.81
& +0.60 [-2.48, +3.81]
& -0.82 [-3.67, +2.06]
& -1.43 [-3.97, +0.88] \\

\midrule

\multirow{3}{*}{16}
& Uniform
& 52.21
& +2.00 [-0.59, +4.80]
& ---
& +1.56 [-2.14, +5.16] \\
& Top-prob
& 50.64
& +0.44 [-3.43, +4.41]
& -1.56 [-5.16, +2.14]
& --- \\
& Probe-guided
& 52.29
& +2.08 [-1.50, +5.87]
& +0.08 [-3.41, +3.78]
& +1.64 [-2.36, +5.73] \\

\midrule

\multirow{3}{*}{32}
& Uniform
& 51.26
& +1.05 [-2.09, +4.30]
& ---
& -0.80 [-3.66, +2.23] \\
& Top-prob
& 52.07
& +1.86 [-1.62, +5.38]
& +0.80 [-2.23, +3.66]
& --- \\
& Probe-guided
& 53.87
& +3.66 [-0.20, +7.62]
& +2.61 [-1.32, +6.68]
& +1.80 [-1.44, +5.20] \\

\midrule

\multirow{3}{*}{64}
& Uniform
& 51.84
& +1.63 [-2.03, +5.30]
& ---
& +1.29 [-3.05, +5.59] \\
& Top-prob
& 50.55
& +0.35 [-3.47, +4.18]
& -1.29 [-5.59, +3.05]
& --- \\
& Probe-guided
& 54.51
& +4.30 [+0.60, +8.28]
& +2.67 [-0.75, +6.22]
& +3.96 [+0.63, +7.62] \\

\bottomrule
\end{tabular}
\caption{
Mean continuation accuracy under online intervention.
Differences are in percentage points; intervals are 95\% paired problem-bootstrap CIs.
Under interventions based on candidate-value readouts, continuation accuracy increases with the budget, with the largest gain at \(B=64\).
}
\label{tab:online_mean_continuation_accuracy}
\end{table*}

\section{Computational Resources}
\label{app:computational-resources}
\textbf{Computational environment.}
Offline probing used 32 logical CPUs and 61.55~GiB of RAM, with Python 3.11.6 and PyTorch 2.12.0.
GPU experiments used eight NVIDIA H20 GPUs, each with approximately 96~GiB of memory.
The GPU environment used Python 3.10.12, PyTorch 2.9.0 (CUDA 12.8), Transformers 4.57.0.dev0, and vLLM 0.11.1.
Answer verification used math-verify 0.5.2.

\textbf{Computational cost.}
We estimate the computational cost of dataset construction, including problem selection and rollout generation.
The estimated costs are 2,755 GPU-hours for RCTV-Core, 7,231 GPU-hours for Qwen3-8B, and 2,569 GPU-hours for DS-Llama-3.1-8B.
Rollout generation for RCTV-InterEnt required approximately 1,224 GPU-hours.

\section{Dataset License}
\label{app:dataset-license}

The source problem pool, Big-Math-RL-Verified, is distributed under the Apache 2.0 license, which permits modification and redistribution subject to its terms.

\section{AI use statement}
We used generative AI tools to help search for relevant literature, assist with coding, and edit manuscript language for readability. All AI-assisted work was reviewed by the authors, and all code changes were manually checked and tested.

\end{document}